\documentclass[lettersize,journal]{IEEEtran}

\usepackage{cancel}
\usepackage{ieee_tra}
\usepackage[most]{tcolorbox}
\usepackage{tabularx}  
\usepackage{pifont}
\usepackage{multirow}
\usepackage{listings}
\newcommand{\ignore}[1]{}
\usepackage{fontawesome}
\usepackage{booktabs}

\usepackage{booktabs}

\definecolor{codegreen}{rgb}{0,0.6,0}
\definecolor{codegray}{rgb}{0.5,0.5,0.5}
\definecolor{codepurple}{rgb}{0.58,0,0.82}
\definecolor{backcolour}{rgb}{0.95,0.95,0.92}
\definecolor{darkgreen}{rgb}{0,0.5,0}

\lstdefinestyle{mystyle}{
    backgroundcolor=\color{backcolour},   
    commentstyle=\color{codegreen},
    keywordstyle=\color{codepurple},
    numberstyle=\tiny\color{codegray},
    stringstyle=\color{codepurple},
    basicstyle=\ttfamily\footnotesize,
    breakatwhitespace=false,         
    breaklines=true,                 
    captionpos=b,                    
    keepspaces=true,                                  
    numbersep=5pt,      
    frame = shadowbox,
    showspaces=false,                
    showstringspaces=false,
    showtabs=false,                  
    tabsize=2
}

\newcommand{\boxedthm}[1]{
\begin{tcolorbox}[colback=orange!5,
                  colframe=black,
                  width=\linewidth,
                  arc=2mm, auto outer arc,
                  boxrule=1pt,
                  boxsep=-1mm,
                 ]
  #1
\end{tcolorbox}
}

\newcommand{\boxedcon}[1]{
\begin{tcolorbox}[colback=gray!5, 
                  colframe=black, 
                  width=\linewidth, 
                  arc=2mm, auto outer arc,
                  boxrule=1pt,
                  boxsep=-1mm,
                 ]
  #1
\end{tcolorbox}
}

\usepackage{balance}
\usepackage{titletoc}  

\begin{document}

\title{{\fontsize{24}{30}\selectfont Joint Depth and Reflectivity Estimation using Single-Photon LiDAR}}

\author{Hashan~K.~Weerasooriya,
        Prateek~Chennuri*,
        Weijian~Zhang*,
        Istvan~Gyongy,
        and~Stanley~H.~Chan
\thanks{Hashan K. Weerasooriya, Prateek Chennuri, Weijian Zhang, and Stanley H. Chan are with the School of Electrical and Computer Engineering, Purdue University, West Lafayette, IN 47906, USA. E-mails: \{hweeraso, pchennur, zhan5056, stanchan\}@purdue.edu. Istvan Gyongy is with the School of Engineering, The University of Edinburgh, Edinburgh, U.K. E-mail: istvan.gyongy@ed.ac.uk. Authors marked with `*' contributed equally to this work.}.  
\thanks{The work is supported in part by the DARPA/SRC CogniSense JUMP2.0 Center, and in part by the National Science Foundation under the grants IIS-2133032 and ECSS-2030570.}
}

\markboth{}%
{Joint Depth and Reflectivity Estimation using Single Photon LiDAR}

\maketitle

\begin{figure*}[b]
    \centering
    \includegraphics[width=1\linewidth, trim= 0mm 00mm 0mm 0mm,clip]{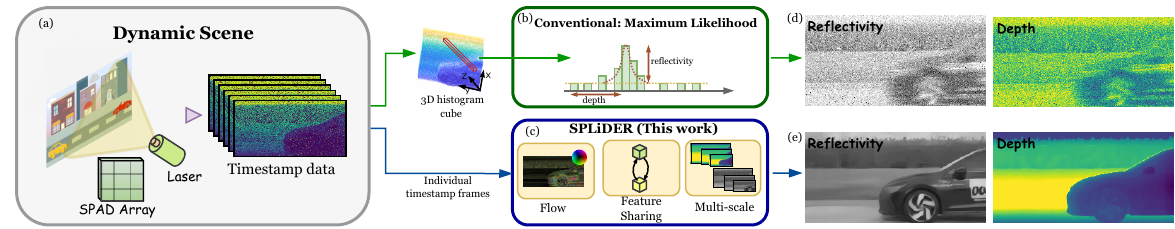}
    \caption{\textbf{Different Processing Methods of Timestamp Data and Corresponding Results:} (a) The SPAD sensor array captures noisy timestamp data from a dynamic scene at a high speed.
    (b) To mitigate noise in raw data, a conventional approach involves clustering multiple detections to form a $3$D cube. Subsequently, the object's reflectivity and depth are determined by identifying the height and the location of the peak through algorithms such as maximum likelihood estimation.
    (c) We propose SPLiDER, a deep learning framework that leverages individual timestamp frames.
    (d) Conventional algorithms suffer from blurry results due to a long integration time.
    (e) Our proposed method yields better results.}
    \label{fig: goal_of_this_paper}
\end{figure*}

\begin{abstract}
Single-Photon Light Detection and Ranging (SP-LiDAR) is emerging as a leading technology for long-range, high-precision $\boldsymbol{3}$D vision tasks. In SP-LiDAR, timestamps encode two complementary pieces of information: pulse travel time (depth) and the number of photons reflected by the object (reflectivity). Existing SP-LiDAR reconstruction methods typically recover depth and reflectivity separately or sequentially use one modality to estimate the other. Moreover, the conventional $\boldsymbol{3}$D histogram construction is effective mainly for slow-moving or stationary scenes. In dynamic scenes, however, it is more efficient and effective to directly process the timestamps. In this paper, we introduce an estimation method to \textbf{simultaneously} recover both depth and reflectivity in fast-moving scenes. We offer two contributions: (1) A theoretical analysis demonstrating the mutual correlation between depth and reflectivity and the conditions under which joint estimation becomes beneficial. (2) A novel reconstruction method, ``SPLiDER'', which exploits the shared information to enhance signal recovery. On both synthetic and real SP-LiDAR data, our method outperforms existing approaches, achieving superior joint reconstruction quality. 

The code is available on the \href{https://hashan-kavinga.github.io/SP-LiDeR/}{\textit{project webpage} \faGithub}.

\end{abstract}


\section{Introduction}
\label{sec:intro}

With the rapid growth of $3$D applications from autonomous driving to virtual reality, Single-Photon Light Detection and Ranging (SP-LiDAR) has become one of the most important technologies for range-related products \cite{boretti_2024_prespective_lidar, Morimoto_2020_SPAD, Li_2023_SPAD, Rapp_2020_SPM, McCarthy_2013_kmrange, mora_martin_21_object_detection, Li_2020_LiDARReview}. The sensing principle of SP-LiDAR involves sending a laser pulse train to an object and using a Single-Photon Avalanche Diode (SPAD) \cite{Becker_2005_TCSPC,Charbon_2013_spad-based} in combination with the Time-Correlated Single-Photon Counting (TCSPC) technique \cite{Becker_2005_TCSPC} to record the time it takes for the pulse to travel \cite{Edoardo_2019_modelling, Torben_2023_spad_simulation}.

The measured timestamps of SP-LiDAR contain two pieces of information: (1) Depth—the time it takes for the pulse to travel, indicating how far the object is; and (2) Reflectivity—the number of photons reflected by the object, as shiny objects reflect photons while dull objects absorb them. Recovering both depth and reflectivity from timestamp measurements is of great interest. Typically, for better reconstruction results, conventional algorithms require multiple photon detections—around hundreds per pixel—which leads to slower acquisition speeds and limits applications to static or slow-moving scenes. However, recovering information in fast-changing environments remains an open challenge, where the choice of data acquisition strategy becomes critical to reconstruction quality.

In the literature, SP-LiDAR measurements are typically processed using two approaches as summarized in \cref{fig: goal_of_this_paper}.

\noindent \textbf{1) $\boldsymbol{3}$D Histogram Cube}: Conventionally, when a SPAD detector is employed, timestamp measurements are accumulated over multiple acquisition cycles until the desired photon level is achieved. This is because collecting more photons allows for better statistical estimation of depth and reflectivity by reducing the randomness due to shot noise, background photon detections, pulse width, and jitter time. For efficient representation of the data, a $3$D histogram cube is constructed where each detection increments the corresponding time-of-flight bin count. Subsequent processing algorithms are applied to determine the location of the histogram peak which corresponds to depth, and the height of the histogram peak which represents reflectivity. By incorporating  $3$D reconstruction algorithms \cite{rapp_2017_unmixing, Shin_2015_3D, peng_2020_non_local, Lindell_2018_SIGGRAPH, peng_2023_boosting}, people have demonstrated $3$D scene reconstruction even in cases where measurements are severely affected by scattering media or multiple surfaces \cite{under_water_2023_Maccarone, halimi_2017_joint_underwater, fog_2022_zhang, Shin_2016_multidepth}.

However, constructing a $3$D histogram cube means that we need to collect data for a longer period which has drawbacks when applied to fast-moving objects.  As the object moves, accumulating photon detections over multiple acquisition cycles makes the reconstruction challenging since each SPAD pixel captures timestamps of a different probability distribution. As a result, we will have multi-peak and broader-peak histograms.

\noindent \textbf{2) Individual Timestamp Frame:} In light of the above challenge, storing timestamp data as individual slices after each acquisition cycle is potentially more advantageous. This strategy, previously proposed by Altmann \etal in \cite{altmann_2020_individual}, enables timestamp frames to be captured over shorter integration periods, thereby minimizing effects caused by motion and making them well-suited for high-frame-rate reconstruction of fast-moving scenes.

Under this processing mechanism, each frame captures the first photon arrival at each pixel within a given time window. Under low-flux and high-speed conditions, the acquisition results in each pixel detecting at most one photon, yielding an average detection rate of less than one photon per pixel per frame \cite{altmann_2020_individual}.  Therefore, the amount of scene information captured per frame is limited, requiring more sophisticated algorithm designs for better reconstruction.

\vspace{5pt}

While both $3$D histogram processing and individual \mbox{timestamp} frame processing can, in principle, support the simultaneous recovery of depth and reflectivity, most existing methods treat them separately. These approaches either reconstruct depth and reflectivity independently or sequentially use one modality to estimate the other. However, since depth and reflectivity information is encoded in the timestamp detections, it seems feasible to \textit{simultaneously} estimate both. This brings us to the two contributions of our paper.


\begin{enumerate}
\item \textit{Does recovering depth help recover reflectivity, and vice versa?} To address this, we derive the maximum-likelihood estimators for the joint recovery problem and analyze the Cramer–Rao Lower Bound (CRLB) to theoretically establish conditions where information sharing is beneficial. This forms the foundation for joint estimation.

\item \textit{Can we design an effective neural network for joint depth and reflectivity recovery in fast-moving scenes?} Building on our theoretical and experimental insights, We propose SPLiDER, a dual-channel joint estimation network with a feature-sharing mechanism. To address information loss from scarce photon detections, we align features from neighboring timestamp frames using optical flow. Our method extracts multiscale features for depth and reflectivity through two parallel branches, enhancing each modality and capturing both fine and coarse details for improved performance.
\end{enumerate}

\section{Prior Work}
\label{sec: prior work}
\noindent \textbf{Depth Estimation}. Depth reconstruction can generally be performed using Maximum-Likelihood Estimation (MLE) and its variants \cite{kirmani_first-photon_2014, Altmann_2016_LiDAR, Tachella_2019_complex, Altman_2016_Bayesian, lee_caspi_2023}. Under the photon-limited regime, where the photon acquisition rate is below $5 \%$ so that dead time becomes negligible \cite{Rapp_2019_Dead, Pediredla_2018_pileup, Gupta_2019_Flooded, zhang_2024_em}, depth estimation is performed by matched-filtering, aligning the photon timestamp registrations with the photon arrival flux function \cite{Bar-David_1969, Chan_2024_CVPR}. In the absence of noise, the matched filter reduces to the sample mean of timestamps when the flux function follows a Gaussian shape. However, while this method performs well with high photon counts, its accuracy degrades under strong background illumination or low photon detections. This is because the matched filter is prone to being trapped in local optima due to its non-convexity. Therefore, pre-processing \cite{kirmani_first-photon_2014, Shin_2015_3D, rapp_2017_unmixing, Yau_2024_MMSP} is required to reject outliers before estimation while leveraging spatial correlations. Advanced depth recovery methods using neural networks are also available \cite{yao_2022_sparsity, zhao_2022_edge_enhancement, zang_2021_non_fusion, ruget_2021_multi_feature, peng_2020_non_local}, often incorporating modalities such as intensity or monocular depth images \cite{Lindell_2018_SIGGRAPH, sun_2020_monocular_fusion, ruget_2021_multi_feature}. All these methods use a $3$D data cube as input.

\noindent \textbf{Reflectivity Estimation}. Reflectivity reconstruction is typically formulated via another MLE of photon counts \cite{kirmani_first-photon_2014,Shin_2015_3D,Altmann_2016_LiDAR, Altman_2017_Binary}, as the reflectivity is proportional to the number of photons reflected by the object. Because of the photon-counting nature, the estimation is formulated by Poisson distribution and/or Binomial distribution for a sum of binary photon detections, similar to quanta image sensors \cite{fossum_2016_quanta, ma_2022_review,chennuri_2024_quiver,Ma_2020_QBP,Liu_2024_bit2bit,chi_2020_dynamic,chan_2016_images,choi_2018_qis,zhang_2024_streamingqis}. Most previous work on reflectivity estimation is based on photon counts rather than the timestamps \cite{rapp_2017_unmixing, vivek_2024_detection}. To avoid exhaustive grid search and the requirement of ground-truth depth in solving the joint MLE problem for reflectivity estimation, an alternative approach \cite{rapp_2017_unmixing} proposes performing noise censoring first and then computing the photon count MLE. Although this method achieves better results, we observe that the information provided by timestamp detections is not fully utilized for reflectivity estimation.

\noindent \textbf{Joint Estimation}. Joint estimation of depth and reflectivity is theoretically beneficial but highly challenging due to the continuous, alternating iterative search between the two parameters. The method presented in \cite{drummond_joint_estimate} adopts a Bayesian formulation, framing joint estimation as a single inference problem, while \cite{halimi_2016_joint_admm, halimi_2017_joint_underwater} jointly estimate depth and reflectivity using the Alternating Direction Method of Multipliers (ADMM) algorithm. To our knowledge, the only deep learning method that considers joint feature extraction from depth and reflectivity is \cite{peng_2023_boosting}, but its decoders are designed independently. 

\noindent \textbf{Dynamic Scenes}.  Depth and reflectivity estimation in dynamic scenes is even more challenging because the events can no longer be aggregated without compensating for motion. While current state-of-the-art methods can handle moving scenes, they rely on multiple timestamp detections per pixel. Consequently, rapid motion relative to the integration time leads to blurred results. Altmann \etal \cite{altmann_2020_individual} propose a depth map reconstruction method using individual photon detection frames, employing a Bayesian approach to model $3$D profile dynamics. An alternative approach is proposed in \cite{gyongy_2020_high_speed}, which uses high-resolution photon-counting frames and multi-event low-resolution timestamp frames in an interleaved manner to achieve high-frame-rate reconstruction. In contrast, our method relies solely on timestamp data.

\section{Depth and Reflectivity Information Sharing}
\label{sec: Mutual Information Sharing}

Most Single-Photon LiDAR (SP-LiDAR) reconstruction algorithms focus solely on depth reconstruction or first decode reflectivity before using it to infer depth. However, since a timestamp data encodes both depth and reflectivity, it would seem more beneficial to reconstruct them simultaneously. The only existing deep learning model for joint depth and reflectivity reconstruction employs two separate decoders to extract depth and reflectivity features independently \cite{peng_2023_boosting}. However, this strict separation may be sub-optimal, as depth and reflectivity features are inherently interconnected.

To that end, this section first theoretically examines how and when the dependency between depth and reflectivity manifests within the per-pixel Maximum-Likelihood Estimation (MLE) framework. Second, we illustrate the shared feature sets between depth and reflectivity using a toy problem. The insights gained from these analyses and the experiments will be incorporated into our proposed algorithm to enhance the feature-sharing mechanism. Proofs of theorems and corollaries are provided in the Appendix and Supplementary Materials.

\subsection{Notation}
\label{subsec: notation}
The notations of this paper follow the literature, e.g., \cite{Shin_2015_3D,rapp_2017_unmixing,Chan_2024_CVPR}. We assume that the SP-LiDAR operates in the low-flux regime, where the dead time effect is negligible. We also assume that there exists a single reflection per pixel with no depth ambiguity and that the object is quasi-static within the per-frame exposure time. Then, the photon arrival can be modeled as an inhomogeneous Poisson process with a mean rate \cite{Bar-David_1969,Snyder_1991_book}
\begin{equation}
    \lambda_{i,j}(\ell, t) = \eta \alpha_{i,j}(\ell) s\left(t - \frac{2 z_{i,j}(\ell)}{c}\right) + b_{\lambda}(\ell),
    \label{eq: reflected pulse}
\end{equation}
where $c$ is the speed of light, $\tau_{i,j}(\ell)=2z_{i,j}(\ell)/c$ is the time delay that carries the true depth $z_{i,j}$, and $s(t) = S \cdot \mathcal{N}(t; 0, \sigma_t^2)$ is a Gaussian-shaped pulse with energy $S$. 

The reflectivity of the object is described by the parameter $\alpha$. The uniformly distributed parameter $b_{\lambda}(\ell) = \eta\lambda_b(\ell)+\lambda_d(\ell) \sim \mathcal{U}(0, t_r)$ is the background noise which contains the ambient light $\lambda_b(\ell)$ and the dark current $\lambda_d(\ell)$, with $\eta$ being the quantum efficiency. To specify the pixel index, we use $(i,j)$. The index $\ell$ is the rank of the frame. \cref{table: Notations} provides a summary of these notations. For notational simplicity, we drop the index terms $\ell$ and $(i,j)$.

\begin{table}[h]
\centering
\caption{Notation overview.}
\label{table: Notations}
\resizebox{0.5\textwidth}{!}{
    \begin{tabular}{c c c c} 
        \toprule
        \textbf{Symbols} & \textbf{Meaning} & \textbf{Symbols} & \textbf{Meaning} \\ \midrule
        $t_r$ & repetition period & $s(\cdot)$ & pulse shape\\
        $N_r$ & \# repetition / frame & $\sigma_t$ & pulse width\\
        $z$ & ground truth depth & $\lambda_b$ & background rate \\
        $\tau =2z/c$ & true time delay & $\eta$ & quant. eff. \\
        $\alpha$ & true reflectivity & $\lambda_d$ & dark current \\
    \bottomrule
    \end{tabular}
}
\end{table}

Given $\lambda(t)$, the core quantity we are interested in is the distribution of the timestamps. Following prior work such as \cite{Bar-David_1969,rapp_2017_unmixing,vivek_2024_detection,Chan_2024_CVPR}, the distribution is defined according to the theorem below.
\boxedthm{
    \begin{theorem}[Joint density of $M$ timestamps $\vt_M$ \cite{Bar-David_1969,vivek_2024_detection}]
    \label{thm: joint density}
    Let $\vt_M = \{t_k\}_{k=1}^M$. For $M \ge 1$,
    \begin{equation}
    p\left[\vt_M, M=m\right] = \frac{e^{-N_r \Lambda(\alpha)}}{m!} \prod_{k=1}^m \ N_r \lambda(t_k; \alpha, \tau).
     \label{eq: joint density}
    \end{equation}
    \end{theorem}
}
In this equation, $\Lambda(\alpha)$ specifies the per-cycle energy of the photon flux, obtained by integrating $\lambda(t)$ over the repetition period $t_r$, via $\Lambda(\alpha) = \int_0^{t_r} \lambda(t) \ dt = \eta \alpha S + B$, where $\eta \alpha S$ is the signal energy and $B = b_{\lambda}t_r$ is the noise energy. The total expected energy per frame is $N_r \Lambda(\alpha)$. We define Signal-to-Background Ratio (SBR) as $\text{SBR} = \eta \alpha S / B$.

The core signal estimation problem is formulated as a joint Constrained Maximum-Likelihood (CML) estimation:
\begin{align}
    (\tauhat , \alphahat) = &\argmax{0 < \tau < t_r, \alpha \geq 0} 
    \Big\{-N_r \eta S \alpha \notag \\
    & + \sum_{k=1}^{m} \log \left( \eta \alpha s \left( t_k - \tau \right) + b_\lambda \right)\Big\},
    \label{eq: joint estimation}
\end{align}
where evidently depth and reflectivity rely on each other as long as $b_\lambda > 0$. In prior works such as \cite{rapp_2017_unmixing}, the joint estimation is solved via two separable problems when one eliminates skeptical outliers and then assumes \emph{zero} background noise:
\begin{corollary}
\label{corrolary: 1}
When $b_{\lambda} = 0$, \cref{eq: joint estimation} simplifies to
\begin{align}
    \alphahat &= \argmax{\alpha \geq 0} \;\Big\{m \log \alpha - N_r \eta \alpha S \Big\} = \frac{m}{N_r \eta S}, 
    \label{eq: marginal reflectivity estimation wo noise} \\
    \tauhat &= \argmax{0 < \tau < t_r} \; \Big\{\sum_{k=1}^{m} \log \left(s\left( t_k - \tau \right)\right)\Big\} = \frac{1}{m} \sum_{k=1}^m t_k, 
    \label{eq: marginal depth estimation wo noise}
\end{align}
meaning that they can be solved separately.
\end{corollary}
\boxedcon{
\cref{corrolary: 1} states that when $\alpha$ and $\tau$ are separable (i.e., $b_{\lambda} = 0$), recovering one will \emph{not} theoretically offer any help in recovering the other.
}

\subsection{Information sharing under MLE}
\label{subsec: MLE CRLB}
In this subsection, we discuss the theoretical intuition behind how depth and reflectivity can help each other when $b_{\lambda} \neq 0$.

\begin{figure}[t]
    \centering
    \subfloat[Depth estimation results with and without prior knowledge of reflectivity. When reflectivity is unknown, we use the mean value of timestamps (\cref{eq: marginal depth estimation wo noise}). When reflectivity is known, we solve \cref{eq: joint estimation}.]{
        \includegraphics[width=0.96\linewidth]{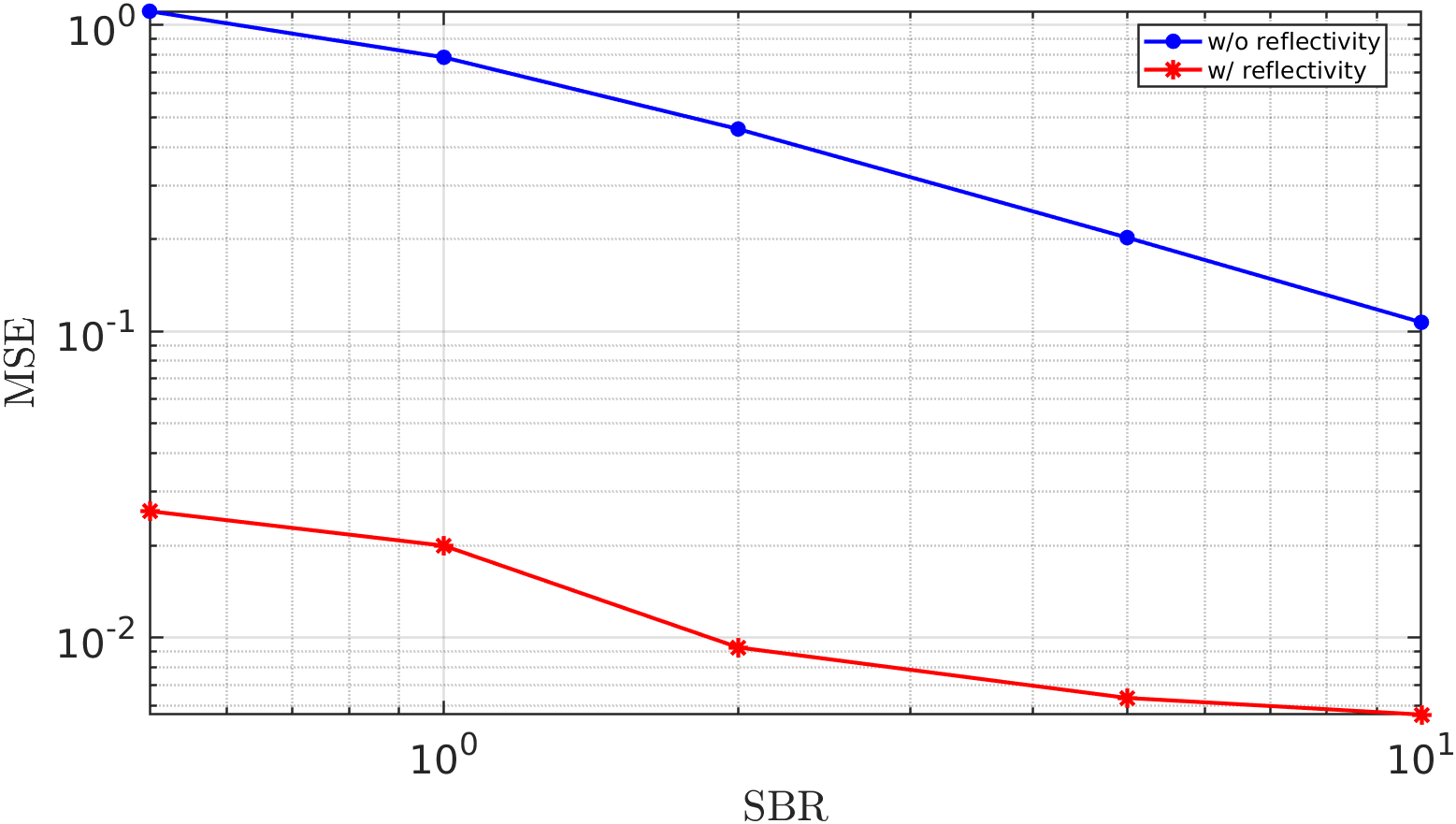}
        \label{fig: depth_estimation}
    }
    \hfill
    \subfloat[Reflectivity estimation results with and without prior knowledge of depth. When depth is unknown, we use \cref{eq: Poisson count CML} to obtain reflectivity. When depth is known, we solve \cref{eq: joint est}.]{
        \includegraphics[width=0.96\linewidth]{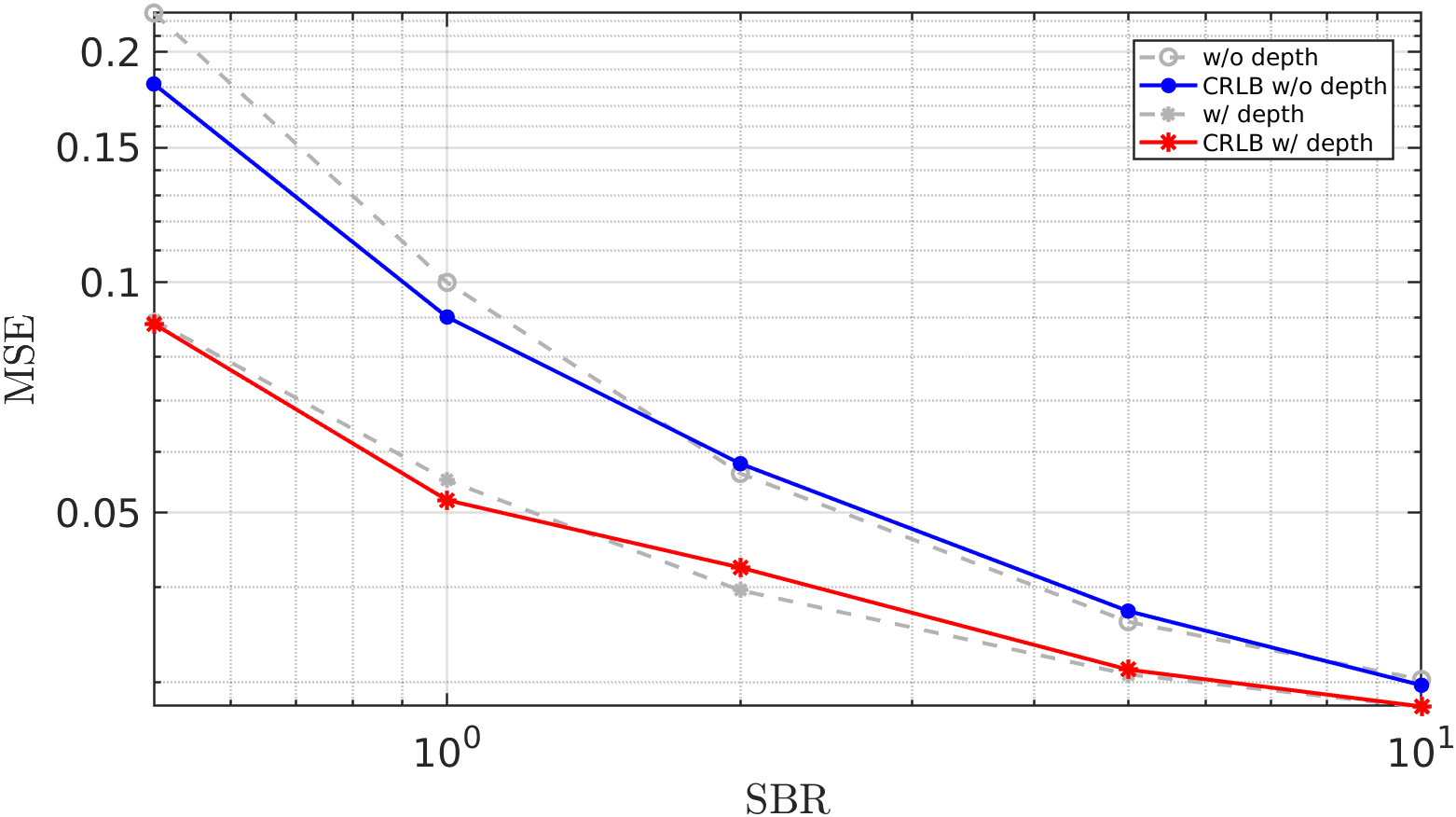}
        \label{fig: reflectivity_estimation}
    }
    \caption{\textbf{Performance Analysis of MLE.} The accuracy of various maximum likelihood estimations under the per-pixel regime across varying signal-to-background ratios. The performance gap increases as noise becomes more dominant.}
    \label{fig: Information_sharing_MLE}
    \vspace{-10pt}
\end{figure}

\vspace{3pt}
\noindent \textbf{Does reflectivity help depth?} This direction is straightforward because if we know reflectivity $\alpha$, we can simplify \cref{eq: joint estimation} to obtain a noisy version of \cref{eq: marginal depth estimation wo noise}. This means that we can run a log-matched filter estimator similar to \cref{eq: marginal depth estimation wo noise} to estimate the depth $\tauhat$. If $\alpha$ is unknown and if we apply \cref{eq: marginal depth estimation wo noise} directly, we will overlook the effect of $\alpha$ and therefore obtain a worse solution. \cref{fig: depth_estimation} shows that the depth estimator with the knowledge of reflectivity consistently outperforms the one without, demonstrating that reflectivity helps depth estimation.

\noindent \textbf{Does depth help reflectivity?} The reverse is non-trivial, as photon count is typically used to estimate reflectivity, and the additional information embedded in timestamps has not been explored for reflectivity estimation until \cite{vivek_2024_detection}. The former estimation, \cref{eq: Poisson count CML}, which is based on photon counts, does not require depth information, whereas the latter does. We compare these two estimators here.

\noindent \textit{\underline{Reflectivity Estimator w/o depth.}} Since the number of photon counts is a Poisson random variable that only conveys reflectivity information, we can derive a CML estimator of reflectivity (see Supplementary Materials)
\begin{equation}
    \label{eq: Poisson count CML}
    \alphahat_{\text{c}} = \max\biggl\{\underset{\alphahat_{\text{c}}^*}{\underbrace{\frac{1}{\eta S}\left(\frac{m}{N_r} - B\right)}}, \ 0 \biggl\},
\end{equation}
where $\alphahat_{\text{c}}^*$ is an \textbf{unconstrained} MLE for which we can derive the CRLB to analyze its performance.
\begin{corollary}
\label{corollary: CRLB w/o depth}
The CRLB of the unconstrained MLE $\alphahat_{\text{c}}^*$ is
\begin{equation}
    \label{eq: CRLB Poisson count}
    \Var\left[\alphahat_{\text{c}}^*\right] \geq \frac{\eta S \alpha + B}{N_r \eta^2 S^2} = \frac{1+1/\text{SBR}}{N_r (\eta S / \alpha)}.
\end{equation}
\end{corollary}
\noindent $\maltese$  \emph{Remark: The variance of the estimator improves when we have more data, higher SBR, or larger system energy ($\eta S$).}

\noindent \textit{\underline{Reflectivity Estimator w/ depth.}} The CML estimate depending on depth can be derived from \cref{eq: joint estimation} as (see Supplementary Materials)
\begin{equation*}
    \alphahat_{t} = \max \left\{\alphahat_{t}^*, 0\right\},
\end{equation*}  
where $\alphahat_{t}^*$ is the largest root to the nonlinear equation below
\begin{equation}
    \label{eq: joint est}
    \sum_{k=1}^m \frac{\eta s(t_k - \tau)}{\eta \alphahat_{t}^* 
 s(t_k - \tau) + B/tr} = N_r \eta S.
\end{equation}
To solve \cref{eq: joint est}, the knowledge of $\tau$ is required. More details on the optimization of this equation can be found in the Supplementary Materials. For this estimator, we can derive the CRLB of the \textbf{unconstrained} MLE $\alphahat_{t}^*$.

\begin{corollary}
\label{corollary: CRLB w/ depth}
The CRLB of the unconstrained MLE $\alphahat_{t}^*$ is
\begin{equation}
    \label{eq: CRLB Poisson timestamp}
    \Var\left[\alphahat_{\text{t}}^*\right] \geq \left[N_r \eta^2 \int_0^{t_r} \frac{s^2(t - \tau)}{\eta \alpha s(t - \tau) + b_{\lambda}} \ dt\right]^{-1}.
\end{equation}
\end{corollary}
\boxedthm{
\begin{theorem}[CRLB comparison between the reflectivity estimators]
\label{thm: CRLB comp}
The variance of $\alphahat_{t}^*$ is uniformly lower than that of $\alphahat_{c}^*$, i.e.
\begin{equation}
\underset{\text{w/ depth}}{\underbrace{\left[N_r \eta^2 \int_0^{t_r} \frac{s^2(t - \tau)}{\eta \alpha s(t - \tau) + b_{\lambda}} \ dt\right]^{-1} }}
\leq 
\underset{\text{w/o depth}}{\underbrace{\frac{\eta S \alpha + B}{N_r \eta^2 S^2}}},
\label{eq: CRLB inequality}
\end{equation}
where the equality holds if and only if $b_{\lambda}=0$, implying that they are equivalent when there is no noise.
\end{theorem}
}
The proof of this theorem can be found in the Appendix. \cref{thm: CRLB comp} provides a theoretical justification for why depth can help reflectivity estimation. 
The numerical results in \cref{fig: reflectivity_estimation} confirm our theory. We remark that this result is consistent with those from \cite{vivek_2024_detection}. However, the context of our results is slightly different: In \cite{vivek_2024_detection}, a sequential estimation was developed whereas in our work, we do joint estimation.

In \cref{fig: reflectivity_estimation}, we compare the MSE of $\alphahat_{\text{t}}$ and $\alphahat_{\text{c}}$. It can be seen that $\alphahat_{\text{t}}$ has a lower MSE\footnote{The observed experimental MSE sometimes goes below the CRLB. This occurs because the positivity constraint imposed on $\alpha$ pulls negative estimations to zero, thereby reducing the error relative to the actual ground truth. Due to the randomness of Poisson realizations, the estimation performance can exceed the theoretical limit set by the CRLB under this positivity constraint.}. This confirms that, under the per-pixel MLE regime, incorporating prior depth knowledge improves reflectivity reconstruction accuracy across all signal-to-background ratios, compared to reflectivity estimation based solely on photon counts arising from Poisson sampling.

In summary, we conclude the following:
\boxedcon{When $b_{\lambda} > 0$, depth helps reflectivity, and reflectivity helps depth.}

\subsection{Information Sharing in the Feature Space}
\label{subsec: Toy problem}

\begin{figure}[t]
    \centering
    \includegraphics[width=1\linewidth, trim= 1mm 0mm 8mm 0mm,clip]{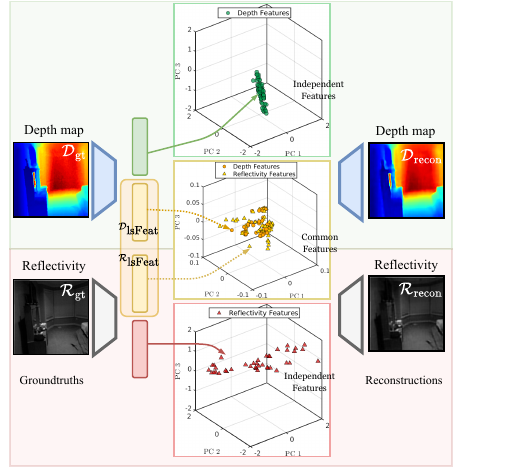}
    \caption{\textbf{Information Sharing Pilot Study.} The setup involves two autoencoders attempting to reconstruct the inputs while isolating the common features in the latent space. The reconstruction results and the corresponding feature distribution verify the claims about shared features. ``PC $1$'', ``PC $2$'' and ``PC $3$'' on the axes represent the $1^{\text{th}}$, $2^{\text{nd}}$ and $3^{\text{rd}}$ principal components.}
    \label{fig: depth_ref_feature_share_experiment}
    \vspace{-15pt}
\end{figure}

Now that we have a theoretical justification for how depth and reflectivity are complementary, we ask the question: \textit{Can this phenomenon be observed even in deep learning implementations?} 

To answer this question, we design a toy experiment to assess the degree of feature overlap between depth and reflectivity. We employ two identical convolutional autoencoders: one to reconstruct reflectivity and the other to reconstruct depth. We simultaneously minimize the distance between specific latent features and the reconstruction losses by using the objective function
\vspace{-3pt}
\begin{equation}
\begin{split}
    \label{eq: loss_func_depth_and_ref}
    \mathcal{L}_{\text{all}} = &\ \mathrm{MSE}(\mathcal{D}_{\text{gt}}, \mathcal{D}_{\text{recon}}) + \mathrm{MSE}(\mathcal{R}_{\text{gt}}, \mathcal{R}_{\text{recon}}) \\
    &\ + \sigma \cdot \mathrm{MSE}(\mathcal{D}_{\text{lsFeat}}, \mathcal{R}_{\text{lsFeat}}).
\end{split}
\end{equation}
For the toy autoencoder network training, we utilized the NYU V2 RGB-D dataset \cite{Silberman_ECCV12_dataset} and assigned a value of $0.5$ to $\sigma$.

\cref{fig: depth_ref_feature_share_experiment} shows an interesting phenomenon of a randomly selected instance. We compare the reconstruction results alongside the corresponding encoded features, visualized in a low-dimensional space. The excellent reconstruction results for both depth and reflectivity indicate that the encoder successfully captures and retains the crucial features needed for accurate reconstruction. The figure further demonstrates that some depth and reflectivity features are clustered in the latent space. This finding aligns with previous work \cite{Lindell_2018_SIGGRAPH, zhang_2018_depth_completion, zhang2023completionformer, Steve_2016_computer_graphics, doCarmo_2016_DifferentialGeometry} and supports the hypothesis that there is a degree of feature sharing between depth and reflectivity. Additional experimental details can be found in the Supplementary Materials.

\section{SPLiDER Network}
\label{sec:method}

\begin{figure*}[t]
    \centering
    \includegraphics[width=1\linewidth, trim= 0mm 0mm 0mm 0mm,clip]{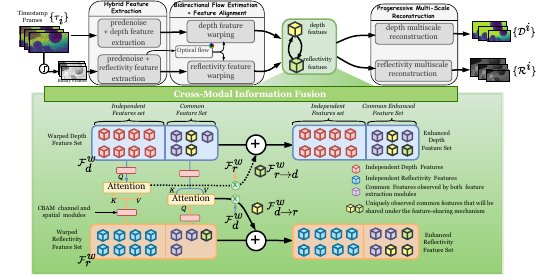}
    \caption{\textbf{Overview of the SPLiDER Network}. The proposed SPLiDER network consists of four main modules. Given several individual timestamp frames directly from a SPAD array, binary frames are generated via a thresholding process: a pixel is marked as 1 if a timestamp exists, and 0 otherwise. Next, depth and reflectivity features from several adjacent input frames are extracted using two feature extraction networks. These features are then warped together based on the optical flow between frames. Subsequently, a cross-model information fusion module is employed to enhance the feature sets of the respective modalities. This module identifies the uniquely observed features by the depth and reflectivity feature extraction mechanisms and fuses them cohesively. Warping and information sharing are performed at three resolution scales to capture both fine and coarse details, improving feature representation. Finally, the multi-scale reconstruction network simultaneously reconstructs depth and reflectivity.} 
    \label{fig: ccam_sp_lider_architecture}
    \vspace{-10pt}
\end{figure*}

This section outlines the proposed SPLiDER  (\textbf{S}ingle \textbf{P}hoton \textbf{Li}DAR joint \textbf{De}pth and \textbf{R}eflectivity). The core component of SPLiDER is Cross-Modal Information Fusion, supported by three additional modules. The overview of the method is shown in \cref{fig: ccam_sp_lider_architecture}.

\subsection{Main Module: Cross-Modal Information Fusion} 

Having established an understanding of the connection between depth and reflectivity, we now address the question of how to effectively implement a feature-sharing mechanism. The starting point is the noisy input timestamp frames, from which noisy binary frames are generated via thresholding. This process does not introduce any new reflectivity information beyond what is already present in the timestamp data.

To ensure that each modality retains its own distinct feature set for accurate reconstruction, we introduce a two-channel estimation procedure for depth and reflectivity. Although noise in the input data may hinder complete feature extraction within each modality, some common features can still be captured by the other. Feature sharing thus helps compensate for missing or incomplete information.

\begin{figure}[b!]
    \vspace{-7pt}
    \centering
    \includegraphics[width=1\linewidth, trim= 0mm 0mm 0mm 2mm,clip]{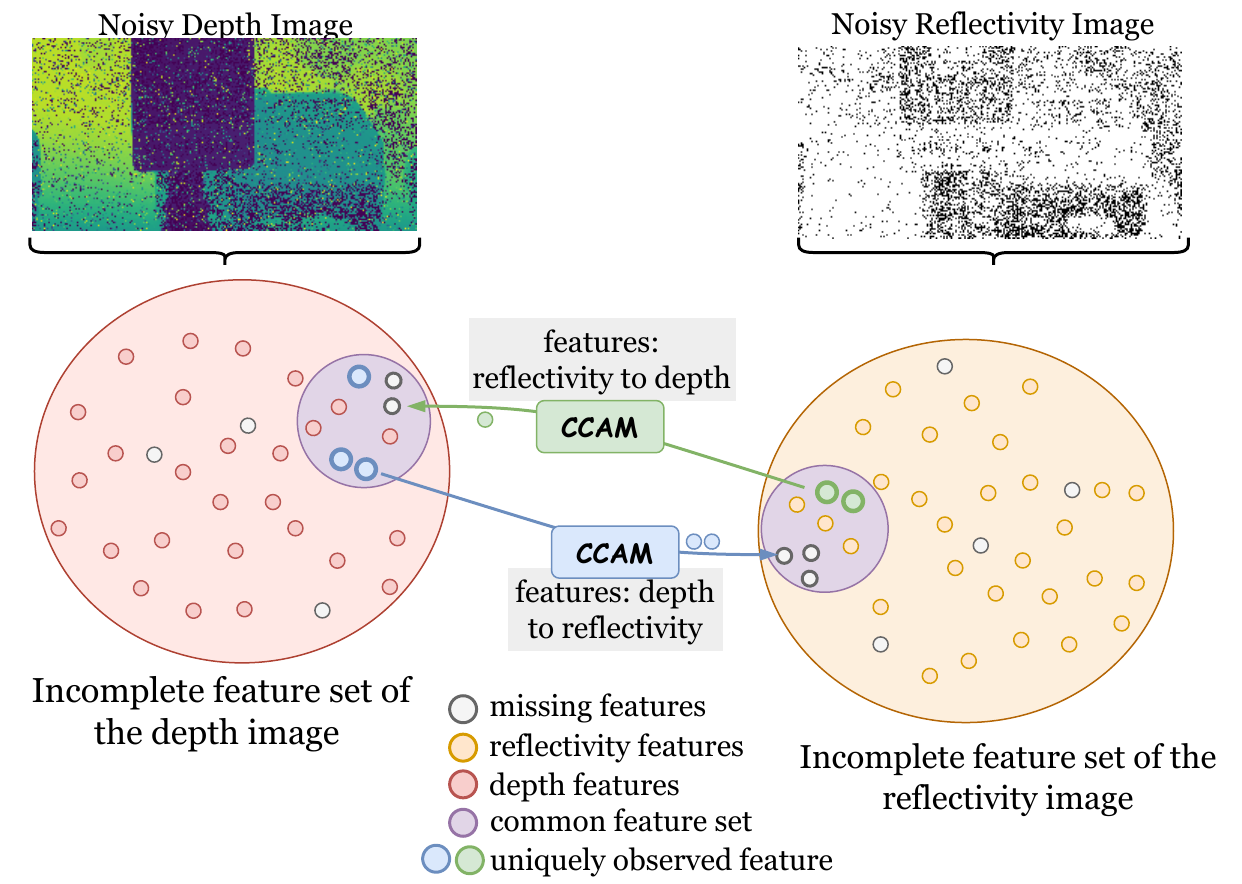}
    \caption{\textbf{Functionality of the CCAM Module.} The CCAM module identifies common features uniquely observed by each channel's feature extraction mechanism and shares them across channels. Due to noisy input, the feature set of the input images is incomplete compared to the feature set of the ground truths.}
    \label{fig: ccam_intuition}
\end{figure}

The proposed two-channel network enables the sharing of common features between modalities through our module, called CCAM (Convolutional Cross-Attention Module). The functionality of the module is illustrated schematically in \cref{fig: ccam_intuition}. Since the timestamp frames are the only unique input, we expect more information to flow from depth to reflectivity than in the reverse direction.

The proposed CCAM module is inspired by Convolutional Block Attention Module (CBAM) \cite{woo_2018_cbam} and the attention mechanism \cite{vaswani_2017_attention}. The main functionalities of these modules are as follows:


\textit{1)} \textit{CBAM: How to extract the features?}
To enable efficient cross-attention, we require a compact summarization of the warped feature maps $\mathcal{F}^w$, as the attention complexity $\mathcal{O}(NM)$ grows rapidly with input lengths $N$ and $M$. To that end, we use CBAM attention modules to generate a channel attention map $\phi_c(\mathcal{F}^w) \in \mathbb{R}^{C\times 1}$ and a spatial attention map $\phi_s(\mathcal{F}^w) \in \mathbb{R}^{H \times W}$:
\begin{align}
    \phi_c(\mathcal{F}) &= \sigma_s(\text{MLP}(\mathcal{F}_{c}^{\text{avg}} + \mathcal{F}_{c}^{\text{max}})) \\
    \phi_s(\mathcal{F}) &= \sigma_s(\text{Conv}([\mathcal{F}_{s}^{\text{avg}}; \mathcal{F}_{s}^{\text{max}}])) 
\end{align}
where $\mathcal{F}_{i}^{\text{avg}}$ and $\mathcal{F}_{i}^{\text{max}}$ for $i \in \{c, s\}$ are the channel or spatial average and max-pooled maps of the warped features $\mathcal{F}^w$, and $\sigma_s$ is the sigmoid function. MLP denotes a multilayer perceptron, while Conv indicates the convolution operation.

This process is repeated to produce $P$ distinct spatial and channel attention maps, which are stacked into unified feature arrays $\Phi_c(\mathcal{F}^w) \in \mathbb{R}^{P \times C}$ and $\Phi_s(\mathcal{F}^w) \in \mathbb{R}^{P \times HW}$. The channel attention module output, $\Phi_c(\mathcal{F}^w)$, identifies salient features (`what') within the input warped frames, while the spatial attention module output, $\Phi_s(\mathcal{F}^w)$, pinpoints their locations (`where'). Together, these modules yield a concise representation of both channel and spatial information that is useful in the feature-sharing mechanism. These channel and spatial attention maps are generated for both depth $\mathcal{F}^w_{d}$ and reflectivity $\mathcal{F}^w_{r}$ warped features independently, yielding $\Phi_s(\mathcal{F}^w_{d})$, $\Phi_s(\mathcal{F}^w_{r})$, $\Phi_c(\mathcal{F}^w_{d})$, and $\Phi_c(\mathcal{F}^w_{r})$.

\textit{2)} \textit{Cross attention: How to share the features?} Cross-attention is then implemented, enabling feature sharing. Cross-attention facilitates the model's ability to learn the inter-dependencies between two input sequences. We explain the feature-sharing mechanism for channel attention maps from reflectivity to depth as follows:
\vspace{-5pt}
\begin{align}
    &\text{Attention}(Q, K, V) = \text{softmax}\left(\frac{QK^T}{\sqrt{d_k}}\right)V, \notag \\
    &\text{head}_c^n = \text{Attention}(Q_c^d W_{n,c}^{Q}, K_c^r W_{n,c}^{K}, V_c^r W_{n,c}^{V}), \notag \\
    &\text{MHA}_c^d(Q_c^d, K_c^r, V_c^r) = \text{Concat}(\text{head}_c^1, \dots, \text{head}_c^P) W^o_c,
\end{align}
where $Q_c^d = \Phi_c(\mathcal{F}^w_{d})$, $K_c^r = V_c^r = \Phi_c(\mathcal{F}^w_{r})$, and $d_k$ denotes the size of $Q$ or $K$. Cross-attention uses depth data ($Q_c^d$: queries) to identify important parts of the reflectivity data ($K_c^r$: keys). It then outputs a weighted version of the reflectivity data ($V_c^r$: values). $W_{n,c}^Q$, $W_{n,c}^K$, and $W_{n,c}^V$ are learnable weight matrices for the $n^{\text{th}}$ attention head of channel features $c$, used to linearly project the query, key, and value before computing attention. We employ a single attention head, which proves sufficient for the feature-sharing mechanism. Spatial cross-attention is defined similarly.

\begin{figure}[t]
    \vspace{-10pt}
    \centering
    \includegraphics[width=1\linewidth, trim= 0mm 0mm 12mm 0,clip]{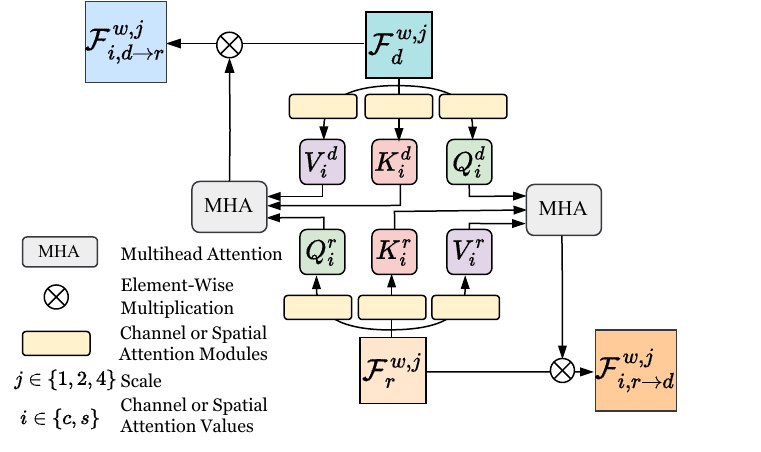}
    \caption{\textbf{Convolutional Cross Attention Module} (CCAM) demonstrates how warped features are used to identify the most relevant features that can be shared across depth and reflectivity channels through an attention head.}
    \label{fig: feature_sharing_module}
    \vspace{-10pt}
\end{figure}

The depth-information-infused reflectivity features for channel attention maps, denoted as $\mathcal{F}^w_{c, r \rightarrow d}$, are produced by the element-wise multiplication of the cross-attention maps with the warped reflectivity features. Similarly, the reflectivity-information-infused depth features for channel attention maps, $\mathcal{F}^w_{c, d \rightarrow r}$, are generated through the same process. $\mathcal{F}^w_{s, d \rightarrow r}$ and $\mathcal{F}^w_{s, r \rightarrow d}$ are similarly defined for spatial attention maps. The generation of these infused feature maps is illustrated in \cref{fig: feature_sharing_module}.

Eventually, to obtain the overall infused feature maps from depth to reflectivity and reflectivity to depth, we use
\begin{align}
\mathcal{F}^w_{d \rightarrow r} &= \sigma (\mathcal{F}^w_{s, d \rightarrow r} + \mathcal{F}^w_{c, d \rightarrow r}) \\
\mathcal{F}^w_{r \rightarrow d} &= \sigma (\mathcal{F}^w_{s, r \rightarrow d} + \mathcal{F}^w_{c, r \rightarrow d})
\end{align}
as specified in \cite{woo_2018_cbam}. This mechanism enables cross-modality fusion, enhancing the original feature map from the respective channel.

\subsection{Supporting Modules}

\subsubsection{\textbf{Hybrid Feature Extraction}} 
SPLiDER simultaneously processes multiple photon timestamp slices, $\{\mathcal{T}_{i}\}_{i=0}^{K}$, where $K$ is the number of input timestamp frames, to enhance the reconstruction quality of the targeted reference frame.

\begin{figure}[t]
    \centering
    \includegraphics[width=1\linewidth, trim= 0mm 0mm 32mm 0mm,clip]{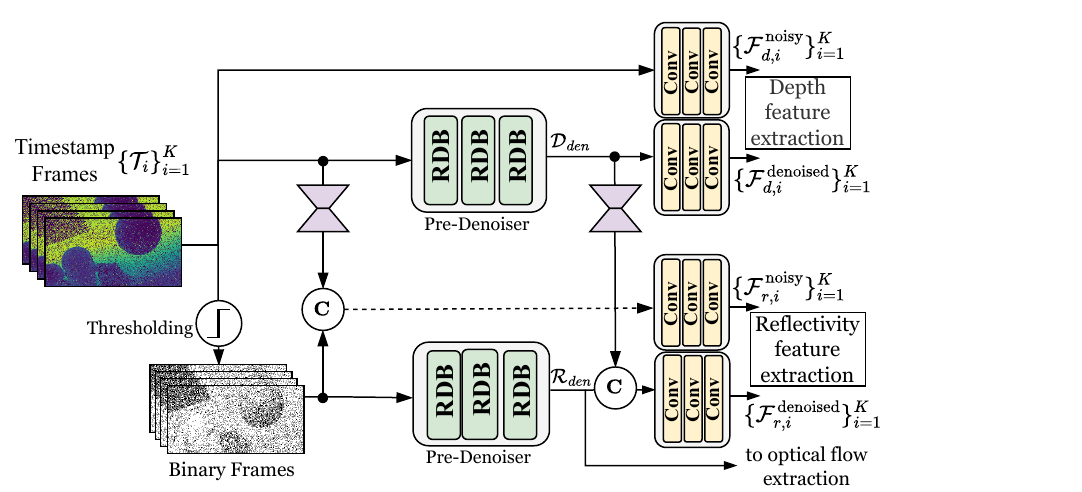}
    \caption{\textbf{Hybrid Feature Extraction Module} denoises timestamp frames and corresponding binary frames separately. Depth and reflectivity features are extracted independently using both denoised and noisy features. Denoised binary frames are used to estimate the optical flow.}
    \label{fig: hybrid_feature_ext_module}
    \vspace{-10pt}    
\end{figure}

Due to noise from dark current, background photons, pulse width, and shot noise, raw data often has a low Signal-to-Noise Ratio (SNR). To improve SNR, we denoise both timestamp frames and binary frames. While denoising removes fine details such as textures, it predominantly preserves the scene structure. By combining high-SNR structural features with fine-grained details from noisy data, we achieve enhanced feature sets $\{\mathcal {F}_{p,i}^{\text{noisy}}\}_{i=1}^{K}$, $\{\mathcal {F}_{p,i}^{\text{denoised}}\}_{i=1}^{K}$, where $p \in \{d, r\}$ represents depth and reflectivity. Depth features are extracted from both denoised and noisy timestamps, whereas reflectivity estimation may require additional cues beyond direct noisy binary frames. Therefore, we use an autoencoder to share coarse feature sets with the reflectivity channel. The overview of the supporting module is depicted in \cref{fig: hybrid_feature_ext_module}.

\subsubsection{\textbf{Bidirectional Flow Estimation $+$ Feature Alignment}} 

\begin{figure}[b]
    \vspace{-12pt}
    \centering
    \includegraphics[width=1\linewidth, trim= 6mm 0mm 12mm 4mm,clip]{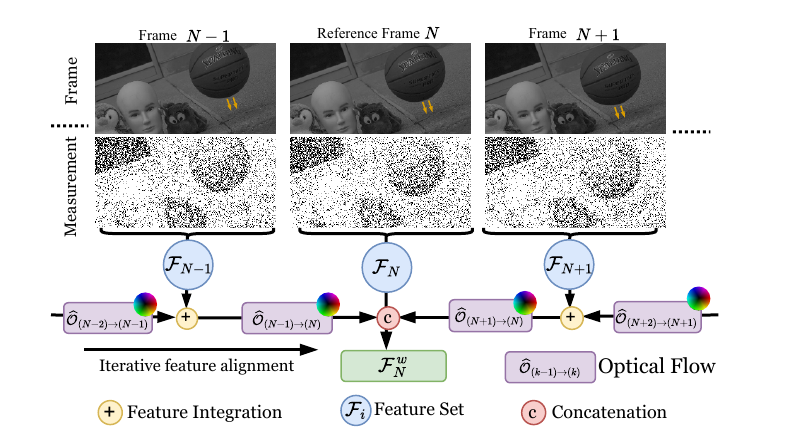}
    \caption{\textbf{Bidirectional Feature Alignment Module} progressively propagates features from both directions to obtain the reference frame $N$ warped features $\mathcal{F}^w_N$.}
    \label{fig: bi-optical_flow_intuition}
\end{figure}

Since the scene is dynamic, simply accumulating neighboring features without accounting for spatial motion across frames is suboptimal. To address this, we propose two modules for flow estimation and temporal data alignment. As illustrated in \cref{fig: bi-optical_flow_intuition}, the core idea of our model is to progressively accumulate features, extracted from the hybrid feature extraction module, from both directions with respect to the reference frame $N$. This warping process is facilitated by an optical flow estimation mechanism, ultimately producing an improved set of features, $\mathcal{F}^{w}_N$, for frame reconstruction.

Inspired by~\cite{zhou2022revisiting}, we design the multi-scale Iterative and Bidirectional Flow Estimation (IBFE) module, which utilizes denoised reflectivity frames for robust flow estimation from noisy sensor data. Additionally, we introduce the Spatial-Temporal Alignment with Residual Refinement (STAR) module, which warps and integrates multi-scale depth and reflectivity features separately, resulting in feature sets $\mathcal{F}^{w,j}_p$, where $j \in \{1, 2, 4\}$ represents the different spatial scales.

\subsubsection{\textbf{Progressive Multi-Scale Reconstruction}} 

Along with the multi-scale reflectivity $\{\mathcal{F}^{w,j}_{r}\}_j$ and depth $\{\mathcal{F}^{w,j}_{d}\}_j$ warped feature maps, we use the multi-scale infused features $\{\mathcal{F}^{w,j}_{d \rightarrow r}\}_j$ for reflectivity estimation and the multi-scale features $\{\mathcal{F}^{w,j}_{r \rightarrow d}\}_j$ for depth estimation.

\begin{figure}[t]
    \centering
    \includegraphics[width=1\linewidth, trim= 17mm 0mm 0mm 0mm,clip]{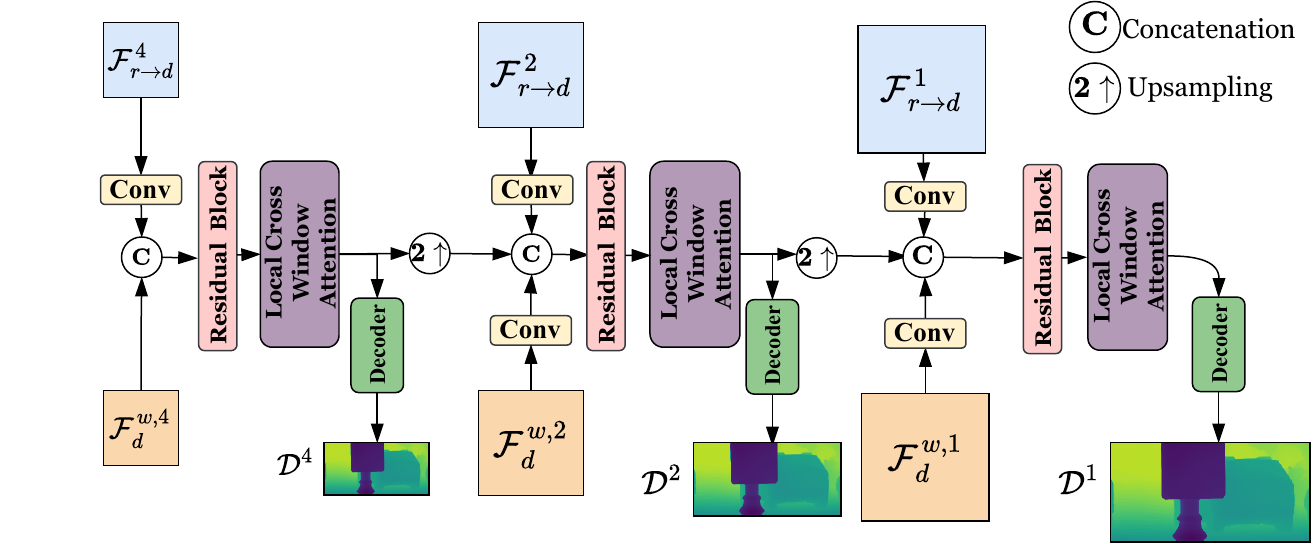}
    \caption{\textbf{Progressive Multi-Scale Reconstruction Module} demonstrates how to reconstruct the multiscale depth maps $\{\mathcal{D}^j\}_j $ for $j=\{1,2,4\}$.}
    \label{fig: multi_scale_reconstruction_module}
    \vspace{-10pt}
\end{figure}

Starting with the lowest scale $j=4$, we combine information from the respective warped features with the infused feature maps using a customized residual network module. Next, the local cross-window attention module~\cite{tu2022maxvit} is applied to generate the output for that scale. The infused feature map is then upscaled and refined at each subsequent scale using the attention and warped feature maps for that scale, repeating this process until the full-scale output is achieved as in \cref{fig: multi_scale_reconstruction_module}.

\vspace{5pt}

\noindent \textbf{Loss Function:} We optimize the weights of the channel branch simultaneously using the following loss function:
\begin{equation}
    \begin{split}
     \mathcal{L}_{\mathcal{H}} = \lambda_1^{\mathcal{H}} \mathcal{L}(\mathcal{H}_{\text{gt}}^1, \mathcal{H}_{\text{den}}) +  \lambda_2^{\mathcal{H}}  \mathcal{L}(\mathcal{H}_{gt}^1, \mathcal{H}^1) + \\ \lambda_3^{\mathcal{H}}  \mathcal{L}(\mathcal{H}_{\text{gt}}^2, \mathcal{H}^2) + \lambda_4^{\mathcal{H}} \mathcal{L}(\mathcal{H}_{\text{gt}}^4, \mathcal{H}^4) + \lambda_5^{\mathcal{H}} \mathcal{P}(\mathcal{H}_{\text{gt}}^1, \mathcal{H}^{1}), 
     \label{eq: overall_loss_function}
     \end{split}
\end{equation}
Where $\mathcal{H}^{j} \in \{\mathcal{D}^{j},\mathcal{R}^{j}\}$ ,respectively, denotes the depth or reflectivity output, while $j$ denotes the scale of the output.  $\lambda_i^{\mathcal{H}}$ is a constant that controls the strength of the regularization for $i =\{1,2,3,4,5\}$. Moreover, $\mathcal{H}_{\text{den}}$ and $\mathcal{H}^j_{\text{gt}}$ refer to the corresponding denoised outputs from the initial denoisers and multiscale ground truth frames, respectively. $\mathcal{P}$ represents the LPIPS loss, and $\mathcal{L}(\mathcal{A}, \mathcal{B})$ is defined as in \cref{eq: indvid_loss_function}.
\begin{equation}
    \begin{split}
   \mathcal{L}(\mathcal{A}, \mathcal{B}) = \|\mathcal{A}-\mathcal{B}\|_1 & \\ + \|\nabla_x \mathcal{A} - \nabla_x \mathcal{B}\|_1 & + \|\nabla_y \mathcal{A} - \nabla_y \mathcal{B}\|_1. 
   \label{eq: indvid_loss_function}
   \end{split}
\end{equation}
Here, $\nabla_x$ and $\nabla_y$ denote the operations of horizontal and vertical gradients. 

\section{Experiments}

In this section, we demonstrate the effectiveness of our proposed method through experiments on both simulated and real-world datasets. We further validate our approach by comparing it against existing reconstruction methods, and provide an ablation study to analyze the impact of key architectural components.

\subsection{Timestamp Simulation Pipeline}
We simulate the first-photon time-correlated imaging mechanism, as detailed in \cite{Henderson_2019_192x128}, for the training and testing of both existing and proposed deep learning methods. This data acquisition mechanism captures only the first photon during each acquisition cycle. As a result, some pixels may receive no photons, leading to an average of less than one photon detection per pixel per frame. 

Existing SP-LiDAR architectures, such as those in \cite{peng_2023_boosting, peng_2020_non_local, Lindell_2018_SIGGRAPH}, typically showcase reconstruction results for static scenes. In such scenarios, it is feasible to fix the signal photon levels (and thus the SBR) by adjusting the scene's reflectance, as signal photon levels are proportional to reflectance while the noise level remains constant. However, maintaining specific SBR values in dynamic scenes is impractical, as it would require altering the reflectance in every frame. This approach does not represent the actual reflectance variations across frames. Therefore, unlike previous methods, we showcase our results using a simulation mechanism that dynamically adjusts the SBR—ranging from $5$ to $10$—based on the reflectance of each frame.

Given that there is a photon detection in the $(i,j)$, i.e., $M = 1$ in \cref{eq: joint density}, the timestamp distribution of the pixel can be modeled as a mixture of distributions \cite{Shin_2015_3D, Chan_2024_CVPR}:
\begin{equation} 
    \begin{split} 
    p[t_{i,j} | M=1] = \frac{\eta \alpha_{i,j} S}{\eta \alpha_{i,j} S + B} \left( \frac{s\left(t_{i,j} - \frac{2z_{i,j}}{c}\right)}{S} \right) + \\ \frac{B}{\eta \alpha_{i,j} S + B} \left( \frac{1}{t_r} \right).
    \label{eq: timestamp_distribution}
    \end{split} 
\end{equation}
We adopt the LiDAR setup specifications from \cite{Scholes_2023_Fundamental} to calculate the parameters $S$, $\alpha_{i,j}$, $\eta$, $B$, and $t_r$.

\begin{figure*}[t]
    \centering
    \includegraphics[width=1\linewidth]{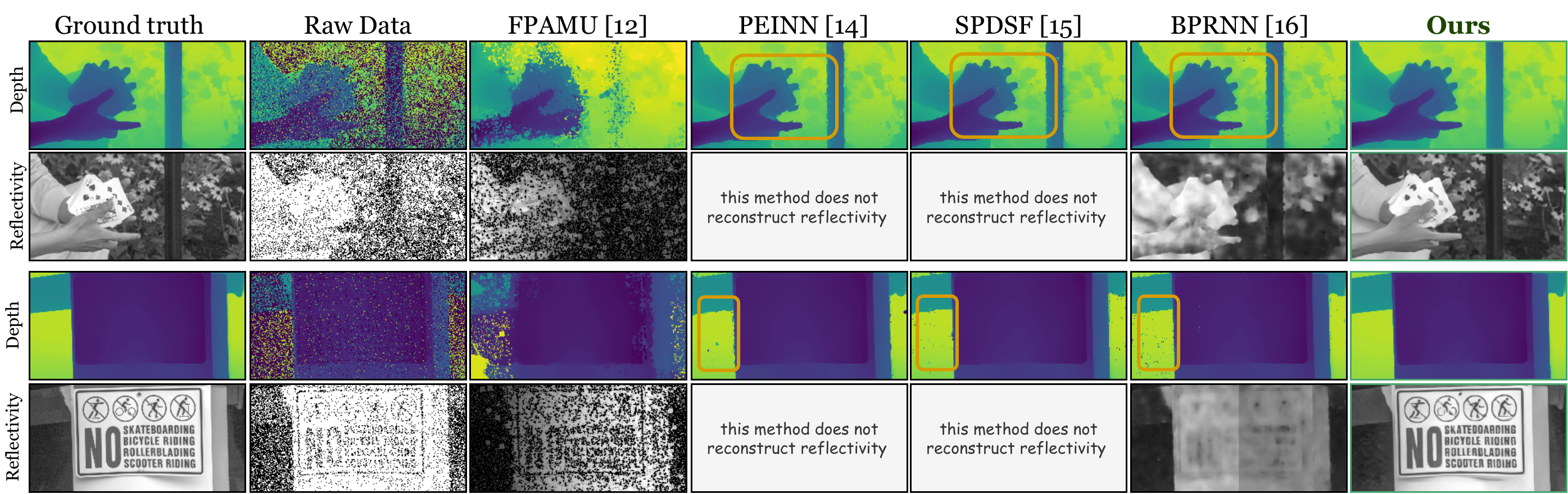}
    \caption{\textbf{Comparison of Depth and Reflectivity Estimation on Simulated Data.} SPLiDER outperforms baseline methods in both reflectivity and depth estimation. The proposed method's robustness to low photon levels effectively suppresses the generation of spurious estimations, a common issue in other approaches (see artifacts in the highlighted regions). SPLiDER yields reflectivity reconstructions with substantially improved detail. \textit{Best viewed in zoom}.}
    \label{fig: simulated_sp_lidar_comparison}
    \vspace{-5pt}
\end{figure*}

Since publicly available depth datasets typically operate within a range of $30$ Hz to $60$ Hz, their temporal resolution is insufficient to replicate the rate of change in object positions captured by single-photon LiDAR sensors, which typically operate at several hundred kHz frame rates. To address this limitation, we use the I2-2000FPS high-speed RGB video dataset~\cite{chennuri_2024_quiver}, which contains $280$ videos. To generate depth maps $z_{i,j}$ from the RGB video dataset, we employ the pre-trained \textit{Depth-Anything v2} network \cite{yang2024depthv2}. Finally, we use the Monte Carlo method to simulate `first-photon behavior' to obtain timestamp frames. Further details of the simulation pipeline are provided in the Supplementary Materials.

\vspace{5pt}
\noindent \textbf{Training of SPLiDER.} We train the network by minimizing the loss function given in \cref{eq: overall_loss_function}, with parameters $\lambda_1^{\mathcal{H}} = 0.2$, $\lambda_2^{\mathcal{H}} = 0.85$, $\lambda_3^{\mathcal{H}} = 0.1$, $\lambda_4^{\mathcal{H}} = 0.05$, $\lambda_5^{\mathcal{D}} = 0$, and $\lambda_5^{\mathcal{R}} = 0.05$. The Adam optimizer \cite{Kingma_2015_adam} is used with an initial learning rate of $10^{-4}$. Every time the loss of the network plateaus, the learning rate is reduced by a factor of $0.5$. Training is carried out on an NVIDIA A$100$ Tensor GPU. A subset of $249$ out of the $280$ depth videos is used for training, leaving $31$ videos for evaluation purposes.

\vspace{-5pt}
\subsection{Synthetic Data Experiments}
The comparative experimental results of the proposed approach and existing algorithms are presented in two parts. For these experiments, all deep learning based methods are retrained from scratch using our timestamp simulation scheme.

\noindent \subsubsection{Comparison with other SP-LiDAR Algorithms} \label{subsubsec: sp_lidar_comparsion} We compare our estimation results quantitatively and qualitatively with other SP-LiDAR reconstruction algorithms. Specifically, we present benchmarking results for five existing algorithms: FPAMU \cite{rapp_2017_unmixing}, PEINN \cite{peng_2020_non_local}, SPDSF \cite{Lindell_2018_SIGGRAPH}, and BPRNN \cite{peng_2023_boosting}. Among these, FPAMU is a non-deep learning algorithm, and BPRNN is the only method that simultaneously estimates both depth and reflectivity. All methods require a $3$D histogram cube (or a cluster of detections per pixel) as the input.

To evaluate existing baseline methods, we use a $3$D histogram cube containing timestamp data collected during one acquisition period. The results are presented in \cref{fig: simulated_sp_lidar_comparison} and \cref{tab: simulated_quantitative_comparison}. We observe that existing deep learning algorithms produce depth maps with spurious estimations (see artifacts in the highlighted regions in \cref{fig: simulated_sp_lidar_comparison}), where the depth estimates at certain pixels are erroneous due to low photon levels—particularly when the level falls below $0.75$ photons per pixel per frame—resulting in less accurate depth reconstruction. However, our method is able to recover the depth maps without any spurious depth estimations. Notably, our SPLiDER surpasses the most recent SP-LiDAR deep learning method, BPRNN, in terms of both reconstruction quality and similarity metrics.

\begin{figure}[b]
    \vspace{-10pt}
    \centering
    \includegraphics[width=1\linewidth]{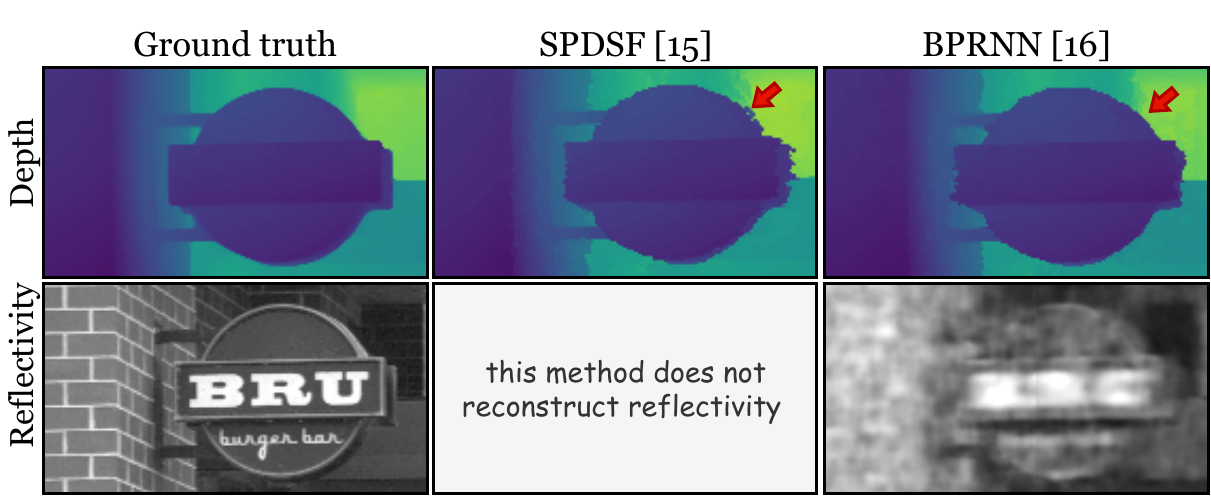}
    \caption{\textbf{The Effect of Longer Acquisition Time for Existing Methods}. When multiple timestamp frames are compiled to construct a $3$D cube, conventional deep learning methods yield blurry reconstruction results. \textit{Best viewed in zoom}.}
    \label{fig: motion_blur}
\end{figure}

Since SPLiDER processes multiple frames, one might wonder how other baseline methods perform when provided with a $3$D histogram cube constructed from timestamp measurements across several acquisition periods. While the increased average number of detections per pixel helps reduce spurious depth estimations, longer acquisition interval introduces blurry reconstruction results due to scene movement, as illustrated in \cref{fig: motion_blur}.

\noindent \subsubsection{Comparison with other Video Reconstruction Algorithms} \label{subsubsec: video_comparsion} We evaluate the reconstruction results of state-of-the-art video denoising/deblurring algorithms, as they can be utilized for reflectivity recovery from noisy timestamp data. For benchmarking, we select MemDeblur \cite{Ji_2022_memdeblur}, Spk2ImgNet \cite{zhao_2021_spike}, RVRT \cite{liang_2022_rvrt}, FloRNN \cite{li_2022_flornn}, and QUIVER \cite{chennuri_2024_quiver}.

\begin{figure*}[t]
    \centering
    \includegraphics[width=1\linewidth]{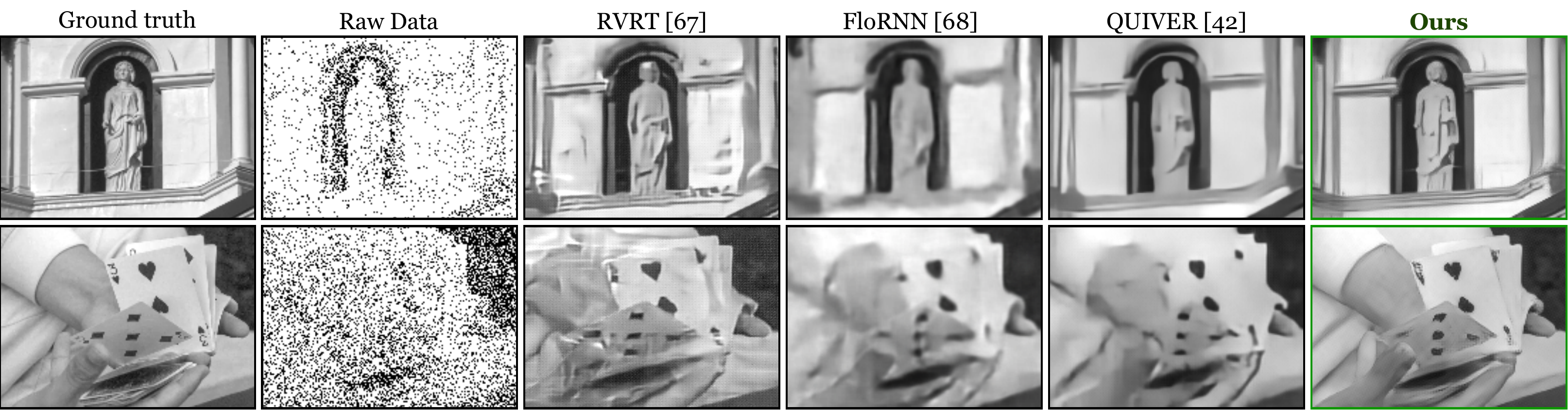}
    \caption{\textbf{Reflectivity Estimation on Simulated Data.}  Visual comparisons with existing video reconstruction algorithms demonstrate better reflectivity from our method, revealing finer details. To ensure a fair comparison, all methods receive $11$ timestamp frames as input.}
    \label{fig: simulated_reflectivity_comparison}
    \vspace{-10pt}
\end{figure*}

\begin{table}[b!]
    \vspace{-10pt}
    \centering
    \caption{Quantitative evaluation of the proposed method against other SP-LiDAR and video reconstruction methods shows that our method achieves superior results. For methods that do not reconstruct depth or reflectivity, the corresponding entries are denoted as `-'. }
    \label{tab: simulated_quantitative_comparison}
    \begin{tabular}{l  c  c  c}
    \hline \multirow{2}{*}{\textbf{Method}} & \multicolumn{2}{c}{\textbf{Reflectivity}} &  {\textbf{Depth}}\\ 
    \cline{2-4}
    & \textbf{PSNR$\uparrow$} & \textbf{SSIM$\uparrow$} & \textbf{RMSE$\downarrow$}\\ 
    \hline
    \multicolumn{4}{c}{\textbf{SP-LiDAR Reconstruction Algorithms}} \\ \hline
    FPAMU (\footnotesize{TCI 17}) \cite{rapp_2017_unmixing} & 9.2556 & 0.0685 & 0.0346 \\ \hline
    PEINN  (\footnotesize{ECCV 20}) \cite{peng_2020_non_local} & - & - & 0.0121\\ \hline
    SPDSF w/ Int (\footnotesize{SIGGRAPH 18}) \cite{Lindell_2018_SIGGRAPH} & - & - & 0.0136 \\ \hline 
    SPDSF w/o Int (\footnotesize{SIGGRAPH 18}) \cite{Lindell_2018_SIGGRAPH} & -  & - & 0.0141 \\ \hline
    BPRNN (\footnotesize{TPAMI 23}) \cite{peng_2023_boosting} & 15.6759 & 0.4171 & 0.0147 \\ \hline
    \multicolumn{4}{c}{\textbf{Video Reconstruction Algorithms}} \\ \hline
    QUIVER (\footnotesize{ECCV 24}) \cite{chennuri_2024_quiver} & 22.1224 & 0.6698 &  -  \\ \hline
    RVRT (\footnotesize{NeurIPS 22}) \cite{liang_2022_rvrt} & 21.8685 & 0.5445 &  -  \\ \hline
    FloRNN  (\footnotesize{ECCV 22}) \cite{li_2022_flornn} & 20.1131 & 0.5675 &   - \\ \hline 
    MemDeblur (\footnotesize{CVPR 22}) \cite{Ji_2022_memdeblur} & 19.8106 & 0.4766 &   - \\ \hline
    Spk2ImgNet (\footnotesize{CVPR 21}) \cite{zhao_2021_spike} & 20.1490 & 0.5722 &  -  \\ \hline 
    \rowcolor[gray]{0.9} \textbf{SPLiDER} & \textbf{23.0260} & \textbf{0.6895} & \textbf{0.0077}  \\ 
    \hline
    \end{tabular}
\end{table}

The results in \cref{fig: simulated_reflectivity_comparison} demonstrate the performance of the different reflectivity reconstruction methods. It is evident that the proposed method yields superior reconstruction results, capturing finer and sharper details that other methods struggle to reproduce. Quantitative results—specifically the PSNR and SSIM values presented in \cref{tab: simulated_quantitative_comparison}—corroborate the observed improvements in reconstruction quality and similarity to the ground truth. Interestingly, the proposed method outperforms the state-of-the-art quanta video restoration algorithm, QUIVER, by more than 0.9 dB in reconstruction quality.
 
\begin{figure*}[b]
    \vspace{-10pt}
    \centering
    \includegraphics[width=1\linewidth]{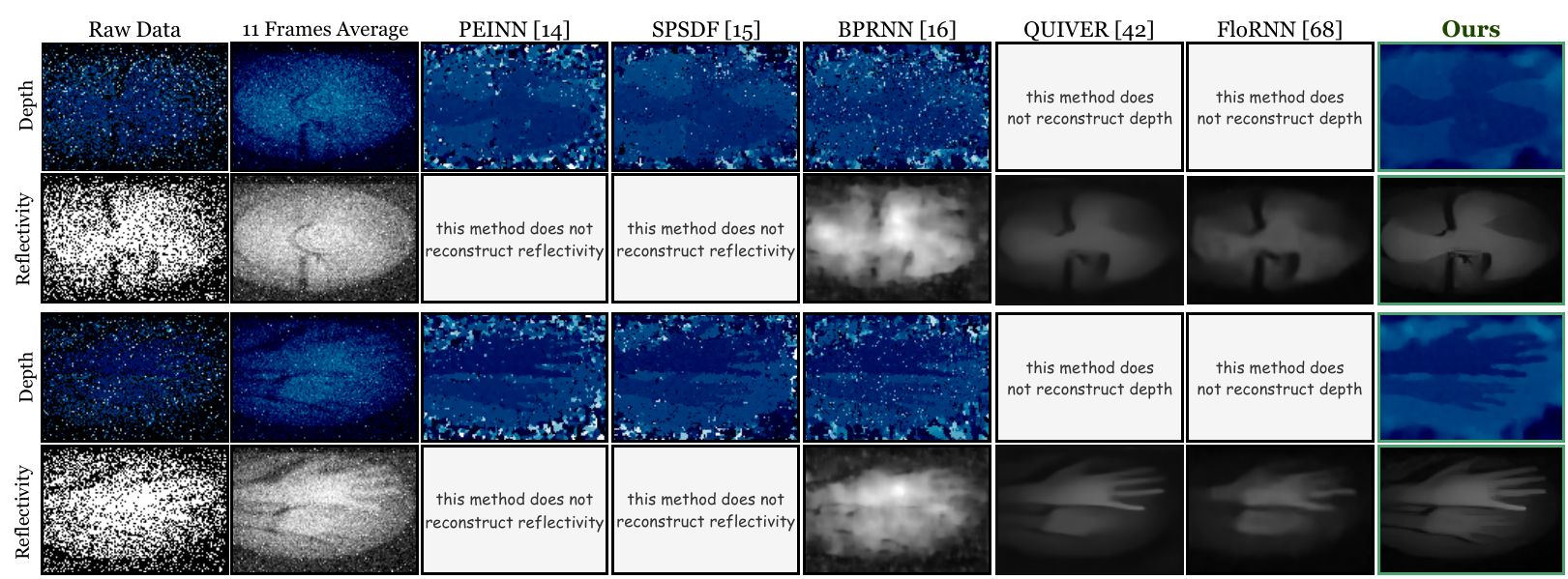}
    \caption{\textbf{Reconstructed Results for Real Data.} Due to the low photon count, existing $3$D reconstruction algorithms often produce depth maps with spurious values—a challenge addressed by our proposed method. Additionally, our approach recovers reflectivity with finer detail compared to existing methods.}
    \label{fig: real_data_results}
\end{figure*}


\subsection{Real Data Experiments}
We capture real timestamp frames using the SPAD sensor detailed in \cite{Henderson_2019_192x128}, which provides a spatial resolution of  $128 \times 192$ and a frame rate of $1000$ frames per second. A Picoquant LDH series $670$ nm picosecond pulsed laser, with $1$ nJ pulse energy, is used to achieve uniform illumination of indoor dynamic scenes having a white background.  The laser operates at $25$ MHz, producing pulses with an effective width of approximately $1$ ns. The system's time-to-digital converter resolution is approximately $35$ ps. 

As shown in \ref{fig: real_data_results}, we compare reconstruction results for two scenarios: a rotating fan and moving hands. Existing $3$D reconstruction techniques, when they are given a histogram cube from a single acquisition cycle, produce depth maps affected by spurious estimations due to low photon levels. In contrast, our proposed method generates precise depth reconstructions. Regarding reflectivity results, existing methods suffer from a loss of reflectivity details. While BPRNN exhibits pronounced smoothing, the other video reconstruction algorithms also struggle to capture sharp details, notably failing to accurately reconstruct the lower hand in the moving-hand scenario. In contrast, our method consistently delivers enhanced reflectivity reconstructions.

\subsection{Ablation}

In our ablation studies, we prioritize examining the effects of the feature-sharing mechanism, as it is the key module, along with the optical flow feature alignment module. To ensure a fair comparison, we increase the dimensions of the feature sets in the initial feature extraction stage to minimize the impact of varying numbers of trainable parameters across different variations.

\begin{figure}[h]
    \centering
    \includegraphics[width=1\linewidth]{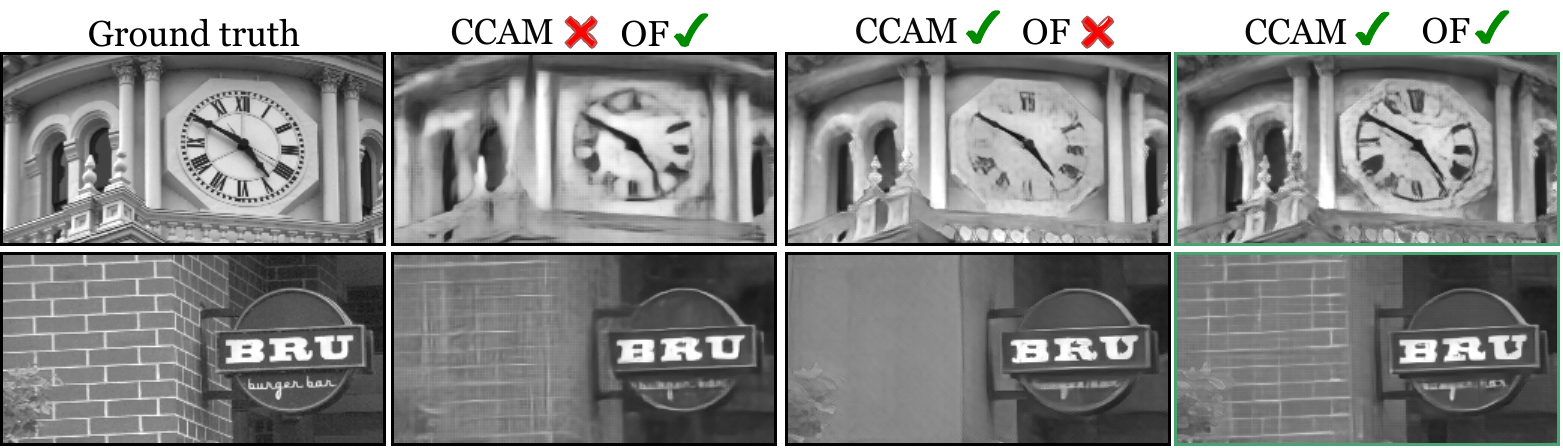}
    \caption{\textbf{Ablation study results.} Visual comparisons of reflectivity estimation illustrate the effectiveness of the CCAM and Optical Flow modules. \textit{Best viewed with zoom}.}
    \label{fig: ablation_results}
    \vspace{-5pt}
\end{figure}

Based on the quantitative comparison presented in \cref{tab: ablation_study_results}, the performance of both depth and reflectivity estimation improves when both modules are included. A comparison of the qualitative results with and without CCAM in \cref{fig: ablation_results} highlights that the feature-sharing module exchanges depth feature sets that complement reflectivity features, leading to improved results. However, under these SBR levels, the effect of the feature-sharing mechanism on depth accuracy is marginal, merely a $2.5 \%$ increase.  This observation aligns with findings from the SPDSF method \cite{peng_2020_non_local}, which shows an average gain of $3.5\%$ from intensity fusion when SBR is above one in low-photon scenarios. Our own comparison in \cref{tab: simulated_quantitative_comparison} further supports this, indicating just a $3.67\%$ accuracy increase with ground-truth reflectivity fusion.

\begin{table}[h]
    \centering
    \caption{Quantitative results of experiments to emphasize the role of optical flow alignment, and feature sharing modules.}
    \renewcommand{\arraystretch}{1.2} 
    \begin{tabular}{>{\centering\arraybackslash}p{1.6cm} >{\centering\arraybackslash}p{1.5cm} c c c}
    \hline 
    \multirow{2}{*}{\textbf{Optical-Flow:OF}} & \multirow{2}{*}{\textbf{CCAM}} & \multicolumn{2}{c}{\textbf{Reflectivity}} & \textbf{Depth} \\ 
    \cline{3-5}
    & & \textbf{PSNR$\uparrow$} & \textbf{SSIM$\uparrow$} & \textbf{RMSE$\downarrow$} \\ 
    \hline
    \ding{55}  & \ding{55} & 21.6954 & 0.6610 & 0.0092 \\ \hline
    \ding{55}  & \ding{51} & 22.2360 & 0.6715 &  0.0084\\ \hline
    \ding{51}  & \ding{55} & 22.0084 & 0.6698 & 0.0079 \\ \hline 
    \rowcolor[gray]{0.9} \ding{51}  & \ding{51} & 23.0260 & 0.6895 & 0.0077 \\ \hline
    \end{tabular}
    \label{tab: ablation_study_results}
\end{table}


\section{Conclusion}
We propose SPLiDER, an end-to-end deep learning framework for the simultaneous reconstruction of depth and reflectivity. By directly processing individual timestamp frames, our method overcomes the limitations of traditional techniques in reconstructing $3$D scenes under dynamic conditions. Unlike independent or single-modality estimation methods, SPLiDER introduces a novel feature-sharing mechanism that enables seamless cross-modal information exchange. We also provide theoretical justification and experimental evidence to demonstrate the complementary relationship between depth and reflectivity. Experiments conducted on both synthetic and real-world datasets demonstrate the superior performance of our approach in producing accurate depth and reflectivity results. An ablation study further highlights the critical role of the feature-sharing module in enhancing reconstruction quality through cross-modal interaction. Overall, SPLiDER provides an effective solution for real-world SP-LiDAR applications, significantly advancing the accuracy of joint depth and reflectivity reconstruction.

\section{Appendix}
In this section, we present detailed derivations of the theorems.

\subsection{Proof of \cref{thm: joint density}}
The goal is to prove that the joint density of the number of photon detections $M$ and the relative timestamps $\vt_M = \{t_k\}_{k=1}^M (0 \leq t_k < t_r)$ during $[0, N_r t_r)$ is
\begin{equation*}
p\left[\vt_M, M=m\right] = \frac{e^{-N_r \Lambda(\alpha)}}{m!} \prod_{k=1}^m \ N_r \lambda(t_k; \alpha, \tau),
\end{equation*}
when $M \geq 1$.

\noindent \textbf{Method 1:}


As mentioned in \cref{subsec: MLE CRLB}, $M$ is a Poisson random variable with a mean $N_r \Lambda(\alpha)$. Thus, its probability mass function (PMF) is
\begin{equation}
    p_M(m; \alpha) = \frac{e^{-N_r \Lambda(\alpha)} \left[N_r \Lambda(\alpha)\right]^m}{m!}, \ m = 0, 1, \ldots
    \label{eq: number_of_photons_poisson}
\end{equation}
Conditioned on $M = m$, the $m$ relative timestamps $\vt_m = \{t_k\}_{k=1}^m, t_k \in [0, t_r)$ are independent and identically distributed according to the normalized photon arrival flux function as in \cref{eq: reflected pulse}, i.e.
\begin{equation}
    f_{{\Tilde{t}_k} | M = m}(t_k | M = m) = \frac{\lambda(t_k; \alpha, \tau)}{\Lambda(\alpha)}, \ t_k \in [0, t_r).
    \label{eq: single_timestamp_stat}
\end{equation}

Therefore, when $M \geq 1$, the joint density is
\begin{align*}
    p\left[\vt_M, M=m\right]
    & = p_M(m) \prod_{k=1}^m \ f_{{\Tilde{t}_k} | M = m }(t_k | M =m) \\
    & = \frac{e^{-N_r \Lambda} (N_r \Lambda)^m}{m!} \prod_{k=1}^m \ \frac{\lambda(t_k)}{\Lambda} \\
    & = \frac{e^{-N_r \Lambda}}{m!} \prod_{k=1}^m \ N_r \lambda(t_k).
\end{align*}
Additionally, when $M = 0$, the joint density reduces to the marginal PMF, i.e. $p\left[\vt_M, M=0\right] = p_M(0) = e^{-N_r \Lambda}$.

\noindent \textbf{Method 2:}


From the prior literature \cite{Bar-David_1969, Chan_2024_CVPR}, if we assume $\vt_M = \{t_j\}_{j=1}^M$ such that $0 \le t_1 < t_2 < \ldots < t_M < t_r$, then for $M \ge 1$,
\begin{equation*}
p\left[\vt_M, M = m \right] = e^{-N_r \Lambda(\alpha)} \prod_{j=1}^m N_r \lambda(t_j).
\end{equation*}

We identify that the statistics are ordered here. If we transform the density of ordered statistics to unordered ones as we need, the density should shrink by $m!$ due to the permutation. The desired result will then emerge.

\subsection{Proof of \cref{thm: CRLB comp}}


The goal is to prove
\begin{equation*}
\left[N_r \eta^2 \int_0^{t_r} \frac{s^2(t - \tau)}{\eta \alpha s(t - \tau) + b_{\lambda}} \ dt\right]^{-1} 
\leq 
\frac{\eta S \alpha + B}{N_r \eta^2 S^2},
\end{equation*}
where the equality holds if and only if $b_{\lambda} = 0$. Note that
\begin{equation*}
    s(t) = S \cdot \frac{1}{\sqrt{2\pi \sigma_t^2}} e^{-\frac{t^2}{2\sigma_t^2}} = S \cdot g(t),
\end{equation*}
where $g(t)$ is defined as the Gaussian probability density function (PDF) and we assume $g(t-\tau)$ is fully supported on the interval $[0, t_r]$, i.e. $\int_0^{t_r} g(t-\tau) \ dt = 1$. Thus, the left-hand side (LHS) becomes
\begin{equation*}
    \text{LHS} = \frac{1}{N_r \eta^2 S^2} \left[\int_0^{t_r} \frac{g^2(t - \tau)}{\eta S \alpha g(t - \tau) + b_{\lambda}} \ dt\right]^{-1}.
\end{equation*}
Then, it is equivalent to proving
\begin{equation*}
    \left[\int_0^{t_r} \frac{g^2(t - \tau)}{\eta S \alpha g(t - \tau) + b_{\lambda}} \ dt\right]^{-1} \leq \eta S \alpha + B.
\end{equation*}

Starting from $\int_0^{t_r} g(t-\tau) \ dt = 1$,
\begin{align*}
    1
    &  = \int_0^{t_r} g(t-\tau) \ dt = \left[\int_0^{t_r} g(t-\tau) \ dt\right]^2 \\
    & = \biggl[\int_0^{t_r} \underset{f(t)}{\underbrace{\frac{g(t-\tau)}{\sqrt{\eta S \alpha g(t - \tau) + b_{\lambda}}}}} \underset{h(t)}{\underbrace{\vphantom{\frac{g(t-\tau)}{\sqrt{\eta S \alpha g(t - \tau) + b_{\lambda}}}} \sqrt{\eta S \alpha g(t - \tau) + b_{\lambda}}}} \ dt \biggl]^2 \\
    & \overset{(i)}{\leq} \int_0^{t_r} \frac{g^2(t-\tau)}{\eta S \alpha g(t - \tau) + b_{\lambda}} \ dt \int_0^{t_r} \left[\eta S \alpha g(t - \tau) + b_{\lambda} \right] \ dt \\
    & = \int_0^{t_r} \frac{g^2(t-\tau)}{\eta S \alpha g(t - \tau) + b_{\lambda}} \ dt \cdot (\eta S \alpha + B),
\end{align*}
where $(i)$ is based on the Cauchy-Schwarz inequality, and the equality holds if and only if $h(t) = k \cdot f(t)$ for $k \neq 0$, or equivalently $b_{\lambda} = 0$.

Rearranging the terms, we obtain
\begin{equation*}
    \left[\int_0^{t_r} \frac{g^2(t - \tau)}{\eta S \alpha g(t - \tau) + b_{\lambda}} \ dt\right]^{-1} \leq \eta S \alpha + B,
\end{equation*}
and the proof is complete.

\twocolumn[
\begin{center}
    {\fontsize{24}{30}\selectfont Supplementary Materials}
\end{center}
\vspace{1em}
]

\startcontents
\printcontents{}{1}{\setcounter{tocdepth}{2}}


\startcontents
\printcontents{}{1}{\setcounter{tocdepth}{2}}

\section{Proofs of Additional Results}
We provide detailed derivations of the equations and corollaries in \cref{sec: Mutual Information Sharing}

\subsection{Joint MLE in \cref{eq: joint estimation}}

We derive the joint constrained Maximum-Likelihood-Estimation (MLE) problem.
\begin{align*}
    (\tauhat , \alphahat)
    & = \argmax{0 < \tau < t_r, \alpha \geq 0} p\left[\vt_M, M=m\right] \\
    & = \argmax{0 < \tau < t_r, \alpha \geq 0} \log \ p\left[\vt_M, M=m\right] \\
    & \overset{(i)}{=} \argmax{0 < \tau < t_r, \alpha \geq 0} \Big\{-N_r (\eta S \alpha + \cancel{B}) - \cancel{\log(m!)} \\
    & \quad\quad\quad\quad\quad\quad\qquad + \cancel{m\log(N_r)} + \sum_{k=1}^{m} \log \lambda(t_k)\Big\} \\
    & \overset{(ii)}{=} \argmax{0 < \tau < t_r, \alpha \geq 0} \Big\{-N_r \eta S \alpha \\
    & \quad\quad\quad\quad\quad\quad\qquad + \sum_{k=1}^{m} \log \left( \eta \alpha s \left( t_k - \tau \right) + b_\lambda \right)\Big\},
\end{align*}
where $(i)$ is based on the substitution of \cref{eq: joint density} and $(ii)$ is based on the elimination of independent terms regarding the optimization variables. 

\subsection{Proof of \cref{corrolary: 1}}


\noindent \textbf{Derivation of \cref{eq: marginal reflectivity estimation wo noise}:}

Assume $b_{\lambda} = 0$ and $\tau$ is a known constant. Then, \cref{eq: joint estimation} becomes
\begin{align*}
    \alphahat
    & = \argmax{\alpha \geq 0} \Big\{-N_r \eta S \alpha + \sum_{k=1}^{m} \log \left(\eta \alpha s \left( t_k - \tau \right)\right)\Big\} \\
    & \overset{(i)}{=} \argmax{\alpha \geq 0} \Big\{m \log \alpha + \cancel{\sum_{k=1}^{m} \log \left(\eta s \left( t_k - \tau \right)\right)} - N_r \eta S \alpha\Big\} \\
    & \overset{(ii)}{=} \argmax{\alpha \geq 0} \underset{L(\alpha)}{\underbrace{\Big\{m \log \alpha - N_r \eta S \alpha\Big\}}},
\end{align*}
where $(i)$ is based on the logarithmic product rule and $(ii)$ is based on the elimination of terms independent of $\alpha$.

To solve $\alphahat$, we take the derivative of $L(\alpha)$ with respect to $\alpha$ and equate it with zero. This will give us
\begin{equation*}
    \frac{d L(\alpha)}{d \alpha} = \frac{m}{\alpha} - N_r \eta S = 0,
\end{equation*}
and thus $\alphahat = m / N_r \eta S$. We remark that the constraint $\alpha \geq 0$ is satisfied because all components are nonnegative.

\noindent \textbf{Derivation of \cref{eq: marginal depth estimation wo noise}:}

Assume $b_{\lambda} = 0$ and $\alpha$ is a known constant. Then, \cref{eq: joint estimation} becomes
\begin{align*}
    \tauhat
    & = \argmax{0 < \tau < t_r} \Big\{-N_r \eta S \alpha + \sum_{k=1}^{m} \log \left[\eta \alpha s \left( t_k - \tau \right)\right]\Big\} \\
    & \overset{(i)}{=} \argmax{0 < \tau < t_r} \Big\{\sum_{k=1}^{m} \log \left[s \left( t_k - \tau \right)\right] - \cancel{N_r \eta S \alpha} + \cancel{m \log (\eta \alpha)}\Big\} \\
    & \overset{(ii)}{=} \argmax{0 < \tau < t_r} \underset{L(\tau)}{\underbrace{\Big\{\sum_{k=1}^{m} \log \left[s \left( t_k - \tau \right)\right]\Big\}}},
\end{align*}
where $(i)$ is based on the logarithmic product rule, $(ii)$ is based on the removal of parameters independent of $\tau$, and $s(\cdot)$ is a Gaussian-shaped function.

To solve $\tauhat$, we take the derivative of $L(\tau)$ with respect to $\tau$ and set it to zero. This will give us
\begin{equation*}
    \frac{d L(\tau)}{d \tau} = \sum_{k=1}^{m} \frac{s \left( t_k - \tau \right) \cdot [(t_k - \tau) / \sigma_t^2]}{s \left( t_k - \tau \right)} = 0,
\end{equation*}
which is equivalent to
\begin{equation*}
    \frac{d L(\tau)}{d \tau} = \sum_{k=1}^{m} (t_k - \tau) = \sum_{k=1}^{m} t_k - m \tau = 0.
\end{equation*}

Therefore, $\tauhat = \frac{1}{m} \sum_{k=1}^{m} t_k$. We observe that the constraint $0 < \tau < t_r$ is automatically satisfied.

\subsection{Reflectivity estimator w/o depth in \cref{eq: Poisson count CML}}


First, we derive an \textbf{unconstrained} MLE from \cref{eq: number_of_photons_poisson}. 
The log-likelihood is
\begin{equation*}
    \log \ p_M(m; \alpha) = -N_r \Lambda(\alpha) + m \log(N_r \Lambda(\alpha)) - \log(m!).
\end{equation*}
To maximize the log-likelihood function, the first-order optimality condition implies that
\begin{equation}
    \frac{\partial \log \ p_M(m; \alpha)}{\partial \alpha} = -N_r \eta S + \frac{m \eta S}{\eta S \alpha + B} = 0.
    \label{eq: supp dPoisson}
\end{equation}
Solving the equation above, the \textbf{unconstrained} MLE becomes
\begin{equation*}
    \alphahat_{\text{c}}^* = \frac{1}{\eta S}\left(\frac{m}{N_r} - B\right).
\end{equation*}
Accordingly, the \textbf{constrained} MLE for the reflectivity without the help of depth is
\begin{equation*}
    \alphahat_{\text{c}} = \max\left\{\alphahat_{\text{c}}^*, \ 0 \right\}.
\end{equation*}

\subsection{Proof of \cref{corollary: CRLB w/o depth}}

\label{subsec: corollary 2 proof}

Based on \cref{eq: supp dPoisson}, the second order derivative is
\begin{align*}
    \frac{\partial^2 \log \ p_M(m; \alpha)}{\partial \alpha^2}
    &= \frac{\partial}{\partial \alpha} \left(-N_r \eta S + \frac{m \eta S}{\eta S \alpha + B}\right) \\
    &= - \frac{m \eta^2 S^2}{(\eta S \alpha + B)^2}.
\end{align*}

The Fisher information is
\begin{align*}
    I(\alpha)
    & \bydef -\E\left[\frac{\partial^2 \log \ p_M(m; \alpha)}{\partial \alpha^2}\right] \\
    & \overset{(i)}{=} \frac{\eta^2 S^2}{(\eta S \alpha + B)^2} \E[m] \\
    & \overset{(ii)}{=} \frac{\eta^2 S^2}{(\eta S \alpha + B)^2} N_r(\eta S \alpha + B) \\
    & = \frac{N_r \eta^2 S^2}{\eta S \alpha + B},
\end{align*}
where $(i)$ is based on the fact that deterministic parameters can be moved out of the expectation, and $(ii)$ follows from the mean of the Poisson random variable $m$ being the total energy over $N_r$ cycles.

Therefore, the Cramer-Rao Lower Bound (CRLB) is
\begin{equation*}
    \Var\left[\alphahat_{\text{c}}^*\right] \geq \frac{1}{I(\alpha)} = \frac{\eta S \alpha + B}{N_r \eta^2 S^2} = \frac{1+1/\text{SBR}}{N_r (\eta S / \alpha)},
\end{equation*}
where Signal-to-Background ratio (SBR) is defined as $\text{SBR} = S/B$.

To the best of our knowledge, although this result matches the CRLB of a Binomial photon count estimator reported in \cite{Shin_2015_3D}, the CRLB derivation for this Poisson photon count reflectivity estimator is new.

\subsection{Reflectivity estimator w/ depth in \cref{eq: joint est}}


Assume that $\tau$ is a known constant. Then, \cref{eq: joint estimation} becomes
\begin{equation*}
    \alphahat = \argmax{\alpha \geq 0} \underset{L_t(\alpha)}{\underbrace{\Big\{-N_r \eta S \alpha + \sum_{k=1}^{m} \log \left( \eta \alpha s \left( t_k - \tau \right) + b_\lambda \right)\Big\}}}.
\end{equation*}
To solve the \textbf{unconstrained} MLE $\alphahat_{\text{t}}^*$, the first order optimality condition implies that
\begin{equation}
    \label{eq: dLt}
    \frac{d L_t(\alpha)}{d \alpha} = -N_r \eta S + \sum_{k=1}^{m} \frac{\eta s \left( t_k - \tau \right)}{\eta \alpha s \left( t_k - \tau \right) + b_\lambda} = 0.
\end{equation}
Rearranging the terms, we obtain \cref{eq: joint est}  as
\begin{equation*}
    \sum_{k=1}^{m} \frac{\eta s \left( t_k - \tau \right)}{\eta \alpha s \left( t_k - \tau \right) + B/t_r} = N_r \eta S.
\end{equation*}
The result is consistent with \cite{rapp_2017_unmixing}. We notice that the left-hand side of \cref{eq: joint est} is monotonically decreasing for $0 \leq \alpha < \infty$. A robust estimation algorithm for $\alpha$ is proposed in~\cref{subsec: robust alg}.

\subsection{Proof of \cref{corollary: CRLB w/ depth}}


Similar to \cref{subsec: corollary 2 proof}, we derive the CRLB for the \textbf{unconstrained} MLE $\alphahat_{t}^*$.

The second-order derivative of the log-likelihood of the joint density is
\begin{align*}
    \frac{\partial^2}{\partial \alpha^2}
    &\left(\log p\left[\vt_M, M=m; \alpha\right]\right)\\
    &= \frac{d}{d\alpha} \left(\frac{d L_t(\alpha)}{d \alpha}\right) \\
    &\overset{(i)}{=} \frac{d}{d\alpha} \left(-N_r \eta S + \sum_{k=1}^{m} \frac{\eta s \left( t_k - \tau \right)}{\eta \alpha s \left( t_k - \tau \right) + b_\lambda}\right) \\
    &= - \sum_{k=1}^{m} \frac{\eta^2 s^2 \left( t_k - \tau \right)}{\left(\eta \alpha s \left( t_k - \tau \right) + b_\lambda\right)^2},
\end{align*}
where $(i)$ is based on the substitution of \cref{eq: dLt}.

The Fisher information is
\begin{align*}
    I(\alpha)
    & = -\E_{M, \Tilde{\vt}_M}\left[\frac{\partial^2}{\partial \alpha^2}\left(\log p\left[\vt_M, M=m; \alpha\right]\right)\right] \\
    & = \E_{M, \Tilde{\vt}_M}\left[\sum_{k=1}^{m} \frac{\eta^2 s^2 \left( t_k - \tau \right)}{\left(\eta \alpha s \left( t_k - \tau \right) + b_\lambda\right)^2}\right] \\
    & \overset{(i)}{=} N_r \Lambda \E_{ \Tilde{t} | M =m }\left[\frac{\eta^2 s^2 \left( t - \tau \right)}{\left(\eta \alpha s \left( t - \tau \right) + b_\lambda\right)^2} \Big| M = m\right] \\
    & \overset{(ii)}{=} N_r \Lambda \int_{-\infty}^{\infty} \frac{\eta^2 s^2 \left( t - \tau \right)}{\left(\eta \alpha s \left( t - \tau \right) + b_\lambda\right)^2} f_{\Tilde{t}| M = m}(t|m) \ dt \\
    & = N_r \Lambda \int_{-\infty}^{\infty} \frac{\eta^2 s^2 \left( t - \tau \right)}{\left(\eta \alpha s \left( t - \tau \right) + b_\lambda\right)^2} \frac{\eta \alpha s \left( t - \tau \right) + b_\lambda}{\Lambda} \ dt \\
    & = N_r \eta^2 \int_{0}^{t_r} \frac{s^2 \left( t - \tau \right)}{\eta \alpha s \left( t - \tau \right) + b_\lambda} \ dt,
\end{align*}
where $(i)$ is due to \cref{lemma: 1} in \cref{subsec: Lemma 1} and $(ii)$ is based on the definition of the expectation of a function of the random variable $t$.

Thus, the CRLB is
\begin{equation*}
    \Var\left[\alphahat_{\text{t}}^*\right] \geq \frac{1}{I(\alpha)} = \left[N_r \eta^2 \int_0^{t_r} \frac{s^2(t - \tau)}{\eta \alpha s(t - \tau) + b_{\lambda}} \ dt\right]^{-1}.
\end{equation*}

Compared to \cite{vivek_2024_detection}, our result provides an explicit formula.

\subsection{\cref{lemma: 1} and its proof}
\label{subsec: Lemma 1}
\begin{lemma}
\label{lemma: 1}
Define $g(t) = \frac{\eta^2 s^2 \left( t - \tau \right)}{\left(\eta \alpha s \left( t - \tau \right) + b_\lambda\right)^2}$. Then,
\begin{equation*}
    \E_{M,\Tilde{\vt}_M}\left[\sum_{k=1}^{m} g(t_k)\right] = N_r \Lambda \E_{\Tilde{\vt}_M | M=m}[g(t) | M = m],
\end{equation*}
where the expectation on the left is taken jointly over $M$ and $\Tilde{\vt}_M$ while the one on the right is conditional on $M=m$ over one single timestamp random variable $t \in [0, t_r)$.
\end{lemma}

\noindent\textbf{Proof of \cref{lemma: 1}}.
\begin{align*}
    & \E_{M,\Tilde{\vt}_M} \left[\sum_{k=1}^{m} \frac{\eta^2 s^2 \left( t_k - \tau \right)}{\left(\eta \alpha s \left( t_k - \tau \right) + b_\lambda\right)^2}\right] \\
    & = \E_{M,\Tilde{\vt}_M} \left[\sum_{k=1}^{m} g(t_k)\right] \\
    & \overset{(i)}{=}  \E_M \left[
    \E_{\Tilde{\vt}_M | M = m} \left[\sum_{k=1}^{m} g(t_k) \Big| M = m\right]
    \right] \\
    & \overset{(ii)}{=} \E_{M} \Bigg[
    \underset{m}{\underbrace{\int_0^{t_r} \ldots \int_0^{t_r}}} \ \sum_{k=1}^{m} g(t_k) \\
    & \qquad\qquad \prod_{k=1}^m \ f_{{\Tilde{t}_k} | M = m}(t_k | m) \ dt_1 \ldots dt_m
    \Bigg] \\
    & = \E_{M} \Bigg[
    \underset{m}{\underbrace{\int_0^{t_r} \ldots \int_0^{t_r}}} \ \left[g(t_1) + \sum_{k=2}^{m} g(t_k) \right] \\
    & \qquad\qquad f_{{\Tilde{t}_1} | M=m}(t_1 | m) \ dt_1 \ldots f_{{\Tilde{t}_m} | M = m}(t_m | m) \ dt_m
    \Bigg] \\
    & \overset{(iii)}{=} \E_{M} \Bigg[
    \underset{m-1}{\underbrace{\int_0^{t_r} \ldots \int_0^{t_r}}} \ \left[\E_{\Tilde{t}_1 | M=m}[g(t_1) | M = m] + \sum_{k=2}^{m} g(t_k) \right] \\
    & \qquad\qquad f_{{\Tilde{t}_2} | M=m}(t_2 | m) \ dt_2 \ldots f_{{\Tilde{t}_m} | M=m}(t_m | m) \ dt_m
    \Bigg] \\
    & \overset{(iv)}{=} \E_{M} \left[ \sum_{k=1}^{m} \E_{\Tilde{t}_k | M=m}[g(t_k) | M = m] \right] \\
    & \overset{(v)}{=} \sum_{m=0}^{\infty} p_M(m) \sum_{k=1}^{m} \E_{{\Tilde{t}} | M=m}[g(t) | M = m] \\
    & \overset{(vi)}{=} \E_{{\Tilde{t}} | M=m}[g(t) | M =m] \sum_{m=0}^{\infty} m p_M(m) \\ 
    & = \E_{{\Tilde{t}} | M=m}[g(t) | M = m] \E_{M}[M] \\
    & = N_r \Lambda \E_{{\Tilde{t}} | M=m}[g(t) | M = m] \\
    & = N_r \Lambda \E_{{\Tilde{t}} | M=m}\left[\frac{\eta^2 s^2 \left( t - \tau \right)}{\left(\eta \alpha s \left( t - \tau \right) + b_\lambda\right)^2} \Big| M = m\right].
\end{align*}
Here, $(i)$ is based on the law of iterated expectation, and the outer expectation is over $M$. $(ii)$ is due to the definition of conditional expectation, and we write the joint density as the product of $m$ marginal densities as in \cref{eq: single_timestamp_stat} due to conditional independence. In $(iii)$, we calculate the integral over $t_1$. In $(iv)$, we calculate the integrals over the other variables and group the results. In $(v)$, we expand the expectation over $m$ and replace the dummy variables $t_k$ ($k = 1, \ldots, m$) by $t$. In $(vi)$, we move the conditional expectation outside the summation. The reason is that the conditional probability density function does not depend on the condition $M = m$ as in \cref{eq: single_timestamp_stat}.

\section{Numerical Verification of Information Sharing under MLE}

In this section, we conduct additional numerical simulations to verify the mutual benefit between depth and reflectivity.
\subsection{Simulation and experimental setups}
We simulate timestamps for one pixel using the Poisson photon arrival statistics under a low photon level and different SBRs. Due to the randomness, we repeat the experiment multiple times. The parameters used in the simulation are summarized in \cref{table: Simulation Specs}.

\begin{table}[h]
\centering
\caption{Simulation specifications. $\{\cdot\}$ means a list of values used.}
\resizebox{0.465\textwidth}{!}{
    \begin{tabular}{c c c} 
        \toprule
        \textbf{Symbols} & \textbf{Meaning} & \textbf{Values} \\ \midrule
        $t_r$ & repetition period & $10$ \\
        $N_r$ & \# repetition & $1000$ \\
        $\tau$ & ground truth time delay & $4$ \\
        $\alpha$ & ground truth reflectivity & $0.5$ \\
        $t$ & temporal duration & $0 \leq t < 10$ \\
        $dt$ & temporal resolution & $1/1000$ \\
        $\sigma_t$ & pulse width & $0.2$ \\
        $N_r \Lambda$ & average photon level & $10$ \\
        SBR & signal-to-background ratio & $\{0.5, 1, 2, 5, 10\}$ \\
        \texttt{iter\_num} & \# repetitive experiments & $\{1000, 50\}$ \\
    \bottomrule
    \end{tabular}
}
\label{table: Simulation Specs}
\end{table}

To verify the information sharing under the Maximum-Likelihood Estimation (MLE) framework, we conduct two parallel experiments to estimate the depth and reflectivity for one pixel from simulated timestamps. One is based on mutual help between depth and reflectivity while the other is not.  We choose $\texttt{iter\_num} = 1000$. The MSE results validate the benefit of information sharing, as can be seen in \cref{fig: Information_sharing_MLE}.

In the case where we use a Gaussian distribution without any background noise to estimate depth and a Poisson distribution to estimate reflectivity, both estimation problems have closed-form solutions, as given by \cref{eq: marginal depth estimation wo noise} and \cref{eq: Poisson count CML}.
However, when $\lambda_b \neq 0$, the two estimators help each other as in \cref{eq: joint estimation}, and no closed-form solution is available. In this case, a search-based algorithm is required. We demonstrate the challenge of local minima in the two optimization problems and propose corresponding solutions in the next two subsections.

\subsection{Depth estimation and a search algorithm}
Assume $\alpha$ is a known constant. Then, \cref{eq: joint estimation} becomes 
\begin{equation}
    \tauhat = \argmax{0 < \tau < t_r} \underset{L_d(\tau)}{\underbrace{\Big\{\sum_{k=1}^{m} \log \left( \eta \alpha s \left( t_k - \tau \right) + b_\lambda \right)\Big\}}}.
    \label{eq: CML_depth}
\end{equation}
To solve the unconstrained MLE $\tauhat_{\text{t}}^*$ embedded in \cref{eq: CML_depth}, it is necessary that
\begin{equation*}
    \frac{d L_d(\tau)}{d \tau} = \sum_{k=1}^{m} \frac{\eta \alpha \dot{s} \left( t_k - \tau \right)}{\eta \alpha s \left( t_k - \tau \right) + b_\lambda} = 0.
\end{equation*}

\begin{figure}[t]
    \centering
    \subfloat[The number of local minima is sparsely located when the SBR is high]{\includegraphics[width=0.4\textwidth]{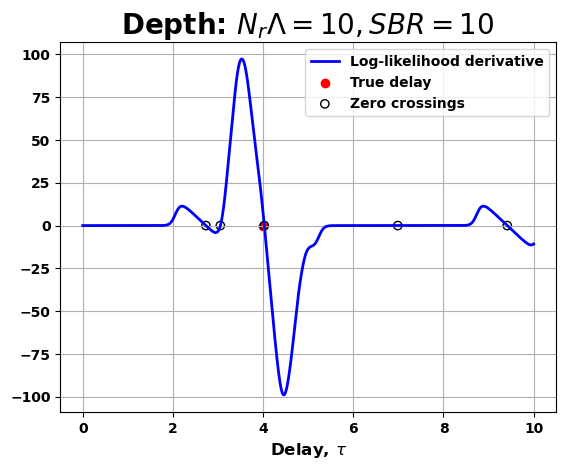}\label{fig: dL high sbr}}
    \hfill
    \subfloat[Low SBR incurs many more local minima near the ground truth.]{\includegraphics[width=0.4\textwidth]{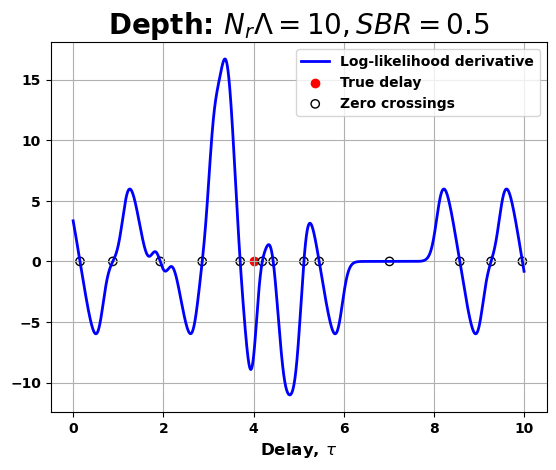}\label{fig: dL low sbr}}
    \caption{Visualization of $\frac{d}{d\tau}L_d(\tau)$ for two different SBR values: one high and one low.}
    \label{fig: dL local minima}
\end{figure}

\begin{figure}[b] 
\centering 
    \includegraphics[width=1\linewidth]{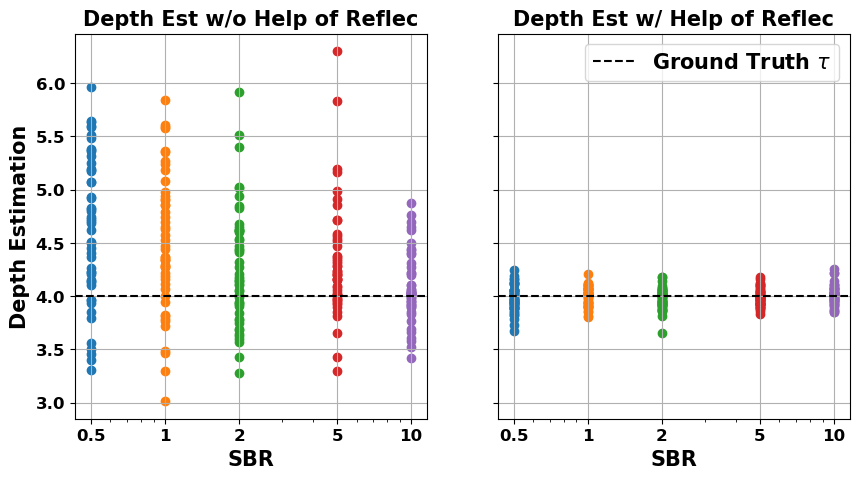} \caption{Depth estimation with and without the help of reflectivity.} 
    \label{fig: dep variance} 
\end{figure}

We note that $\frac{d}{d\tau}L_d(\tau)$ is random because it is a function of random variables $t_k$ $(k = 1, \ldots, m)$. We plot two typical realizations of $\frac{d}{d\tau}L_d(\tau)$ in \cref{fig: dL local minima}. Within the constraint interval $0 < \tau < t_r$, the zero crossings are fewer when the SBR is higher as shown in \cref{fig: dL high sbr} while they are more when the SBR is lower in \cref{fig: dL low sbr}. More severely for the latter case, there are a few zero crossings clustered near the ground truth. Therefore, an ordinary numerical solver tends to be trapped in local minima.

Our goal is to find the zero crossing closest to the ground truth efficiently and robustly. In the proposed \cref{alg:depth_estimation}, we start from a small interval centered at the ground truth and examine the signs of $\frac{d}{d\tau}L_d(\tau)$ at the two boundaries. If they are the same, we gradually increase the interval until the two signs are different. After applying a simple root-finding solver, we find the zero crossing on this interval, and this is the estimated depth. Although the starting point is the ground truth, it is almost impossible for the estimated depth to be the same due to the randomness of $\frac{d}{d\tau}L_d(\tau)$.

\begin{algorithm}
\caption{Robust Depth Estimation Algorithm}
\label{alg:depth_estimation}

\begin{algorithmic}[1]
\Require Initial guess $\tau_0$, step size $\texttt{step}$, maximum iterations $N_{\text{max}}$, iteration number $\texttt{iter}$
\Ensure Estimated depth $\hat{\tau}$

\State Initialize $\hat{\tau} \gets \tau_0$, $a \gets \tau_0$, $b \gets \tau_0$, $\texttt{iter} \gets 1$
\State Define $L_d^{'}(\tau)$ \Comment{Derivative of likelihood function}

\While{sign[$L_d^{'}(a)$] = sign[$L_d^{'}(b)$] \textbf{and} \texttt{iter} $< N_{\text{max}}$}
    \State $a \gets a - \texttt{step}$
    \State $b \gets b + \texttt{step}$
\EndWhile
\If{sign[$L_d^{'}(a)$] $\neq$ sign[$L_d^{'}(b)$]}
    \State Solve $\hat{\tau}$ such that $L_d^{'}(\hat{\tau}) = 0$ using a root-finding method on $[a, b]$
\Else
    \State \textbf{Return} \emph{Error: ``Bracket not found"}
\EndIf

\State \textbf{Return} $\hat{\tau}$
\end{algorithmic}
\end{algorithm}

\noindent \textbf{Depth estimation results. } In addition to \cref{fig: Information_sharing_MLE} from the main text, we provide the scatter plot of $50$ realizations to visualize the estimation performance with and without knowledge of reflectivity, in \cref{fig: dep variance}. According to the figure, with the reflectivity information, the variance of the depth estimation is much smaller, leading to a more accurate estimate.

\subsection{Reflectivity estimation and a robust algorithm}
\label{subsec: robust alg}
\begin{figure}[t]
    \centering
    \subfloat[The trajectory of the first derivative $\frac{d}{d\alpha}L_t(\alpha)$ for a high SBR value of $10$.]{\includegraphics[width=0.4\textwidth]{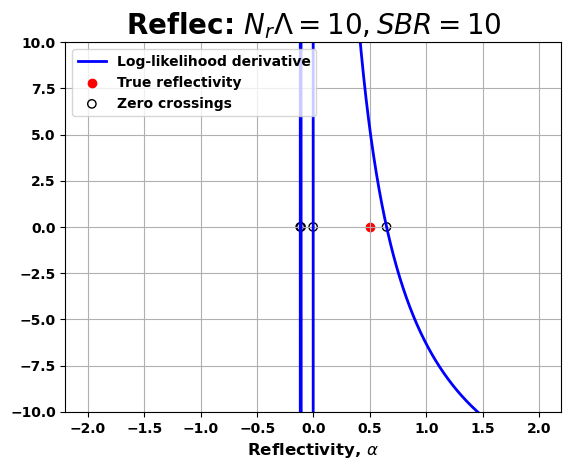}\label{fig: ref local minima high sbr}}
    \hfill
    \subfloat[The trajectory of the first derivative $\frac{d}{d\alpha}L_t(\alpha)$ for a low SBR value of $0.5$.]{\includegraphics[width=0.4\textwidth]{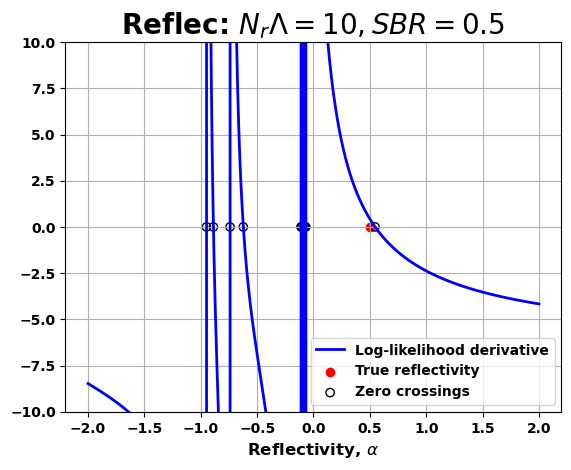}\label{fig: ref local minima low sbr}}
    \caption{Visualization of $\frac{d}{d\alpha}L_t(\alpha)$ for two different SBR values: one high and one low.}
    \label{fig: ref local minima}
\end{figure}

\cref{eq: dLt} is required to solve the reflectivity estimation with the help of depth. Likewise, this optimization problem also suffers from local minima, as illustrated in \cref{fig: ref local minima}.

\begin{figure}[t] 
\centering 
    \includegraphics[width=1\linewidth]{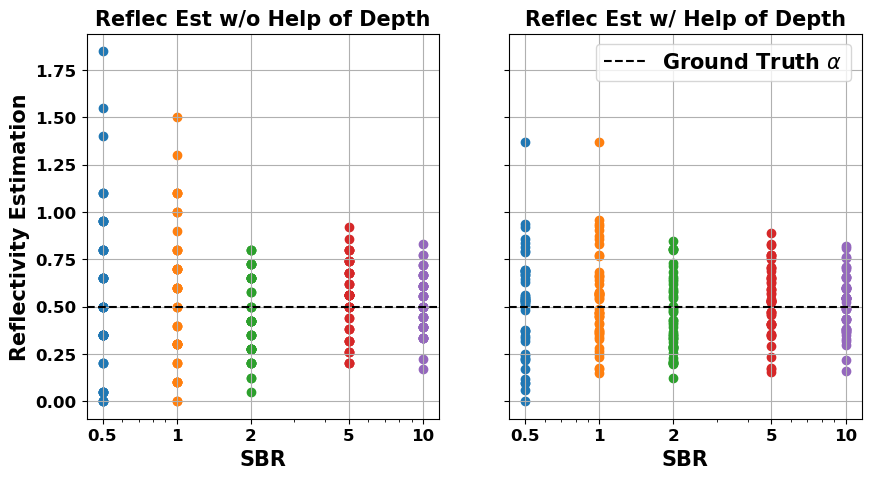} \caption{Reflectivity estimation with and without the help of depth.} 
    \label{fig: reflec variance} 
\end{figure}

The local minimum problem in the reflectivity estimation is more challenging than the depth because the zero crossings in \cref{fig: ref local minima} are even closer to each other compared to those in \cref{fig: dL local minima}. However, we recognize that the $\frac{d}{d\alpha} L_t(\alpha)$ is monotonically decreasing within $[0, \infty)$, and the optimization is also constrained in this region, indicating that the negative zeros should be ignored. Inspired by this, we propose an algorithm based on the bisection method for the reflectivity estimation, as illustrated in \cref{alg:alpha_estimation}.

\begin{algorithm}
\caption{Robust Reflectivity Estimation Algorithm}
\label{alg:alpha_estimation}

\begin{algorithmic}[1]
\Ensure Estimated reflectivity \(\hat{\alpha}\)

\State Initialize \(b\) as a large positive value
\State Define \(L_t^{'}(\alpha)\)  \Comment{Derivative of likelihood function}

\If{\(L_t^{'}(0) \leq 0\)}
    \State \(\hat{\alpha} \gets 0\)
\Else
    \State Solve \(\hat{\alpha}\) such that \(L_t^{'}(\hat{\alpha}) = 0\) using the bisection method on the interval \((0, b)\)
\EndIf

\State \textbf{Return} \(\hat{\alpha}\)

\end{algorithmic}
\end{algorithm}

First, we examine $\left.\frac{d}{d\alpha} L_t(\alpha) \right|_{\alpha = 0}$. If it is not greater than zero, according to the monotonicity of $\frac{d}{d\alpha} L_t(\alpha)$ in $[0, \infty)$, no zero crossing will exist in this region. Therefore, we assign zero to the constrained optimization without any calculations. When $\left.\frac{d}{d\alpha} L_t(\alpha) \right|_{\alpha = 0} > 0$, one unique solution exists in $[0, \infty)$, and we can apply a bisection method to solve it.

\noindent \textbf{Reflectivity estimation results. } Using \cref{alg:alpha_estimation}, we draw the scatter plot to evaluate the reflectivity estimation performance with and without knowledge of depth in \cref{fig: reflec variance}. According to the figure, the reflectivity estimation is much better with the help of depth, resulting in low-variance estimates, particularly when the SBR is low.

\begin{figure}[b] 
\centering 
    \includegraphics[width=1\linewidth]{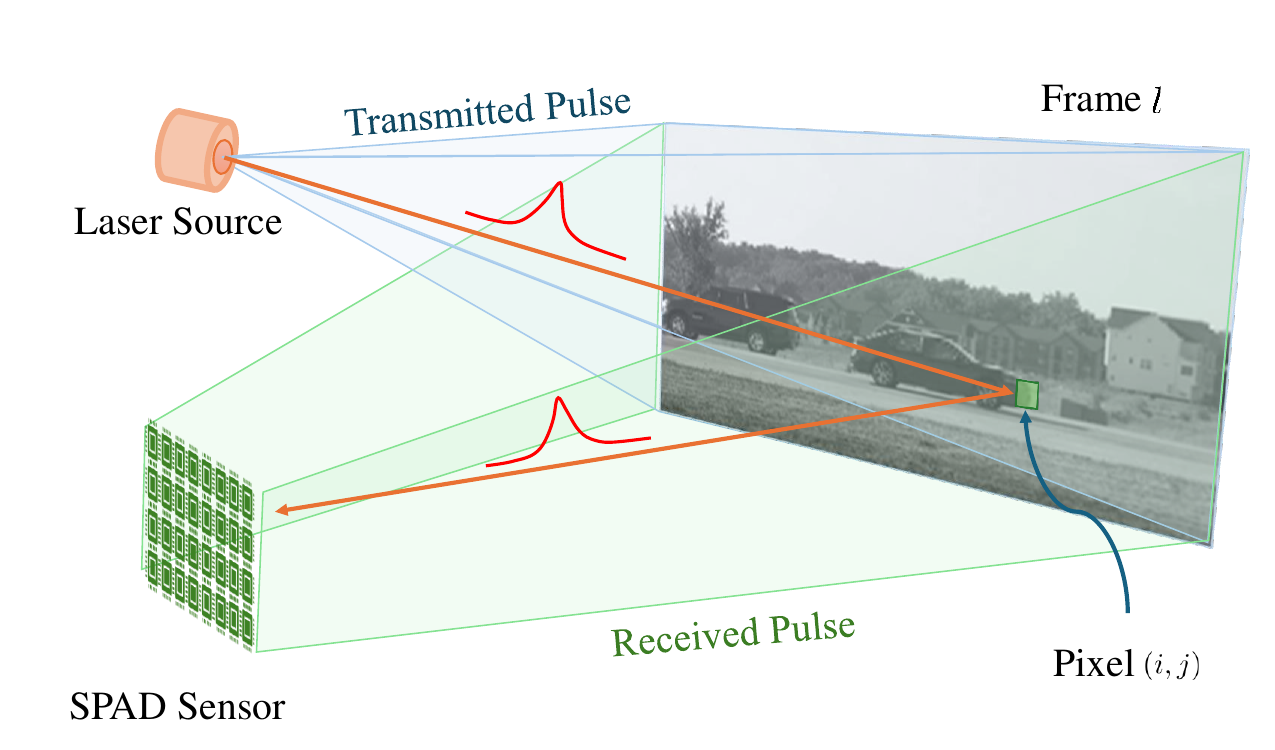} \caption{The single photon LiDAR imaging setup showcases the main components of the arrangement. The laser source emits pulses in rapid succession, whereas the SPAD sensor records the time of arrival of photons.} 
    \label{fig: sp_lidar_imaging} 
\end{figure}

\section{Simulation Pipeline}
\label{sec: Simulation}

In this section, we provide a detailed description of how the SP-LiDAR simulation is conducted. The arrangement of a typical SP-LiDAR setup is illustrated in \cref{fig: sp_lidar_imaging}. As outlined in the manuscript, we adopt typical SP-LiDAR parameter settings, assuming no multi-path interactions, pulse elongation, crosstalk, or dead time. Under these assumptions, we can presume the independence of the captured photons.

\begin{figure*}[t] 
\centering 
    \includegraphics[width=0.97\linewidth]{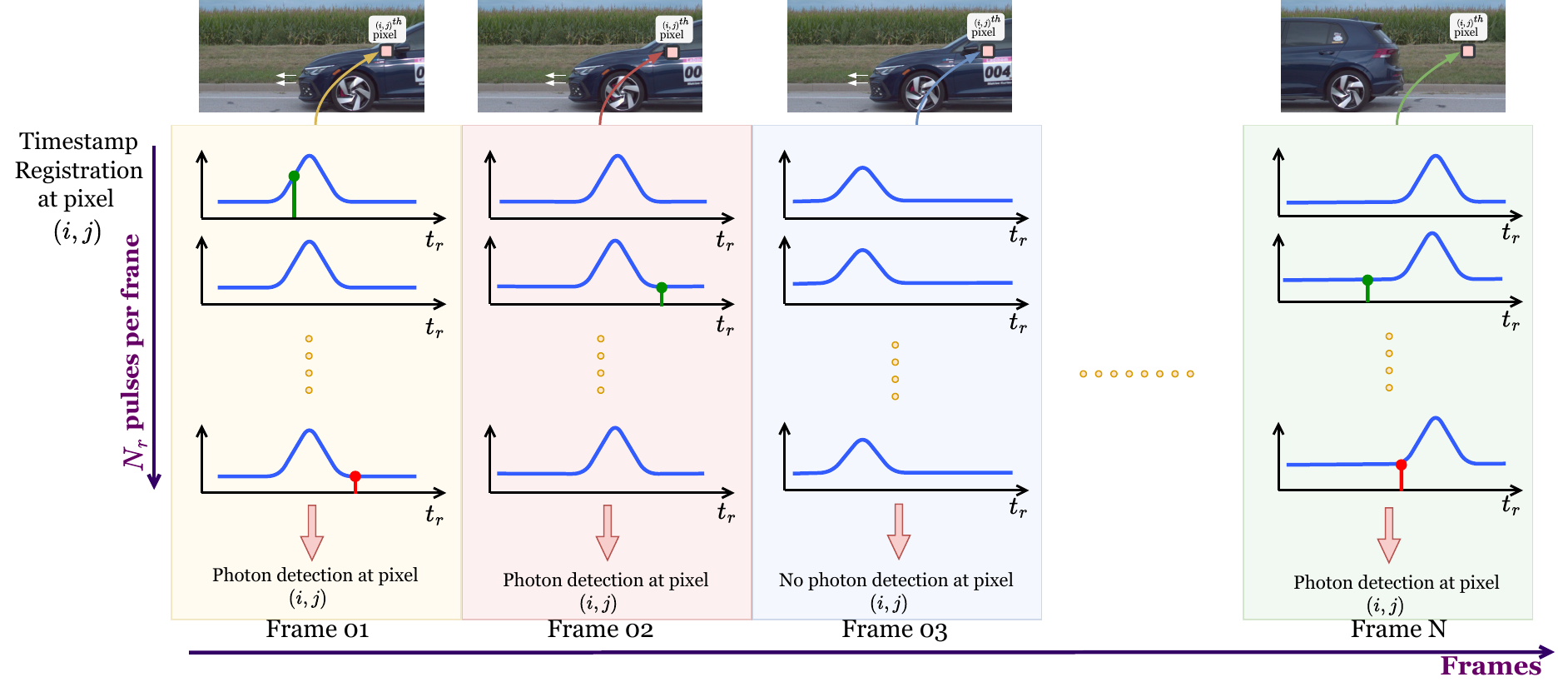} \caption{The timestamp registration of the SPAD sensor captures the behavior of the first photon for the pixel $(i,j)$. During the exposure time $t_{\text{exp}} = N_r t_r$, only the first photon is detected. Successful registrations are marked in green, while missed photons are marked in red. In some frames, there may be no photon registrations due to the random nature of photon arrivals. Timestamps are recorded for all pixels, resulting in the generation of timestamp frames.} 
    \label{fig: first_photon_imaging} 
\end{figure*}

Furthermore, in our simulation, we assume that the scene is quasi-static, meaning it remains stationary during each exposure time. We replicate the operation of the SPAD sensor described in \cite{Henderson_2019_192x128}, which is capable of registering the first photon during each readout cycle, following the first-photon behavior illustrated in \cref{fig: first_photon_imaging}.

The figure illustrates how the first photon arriving at the sensor from the $(i,j)^{\text{th}}$ pixel is captured during each exposure time. Since the laser operates faster than the SPAD sensor’s timestamp readout capability, there are $N_r$ laser repetitions within each exposure. Due to the random nature of photon arrivals, some exposure intervals may result in no detections, while others may register multiple photons. The number of photon detections per exposure time at pixel $(i,j)$ is given by:
\begin{equation}
    p_{M_{i,j}}(m_{i,j}) = \frac{e^{-N_r \Lambda_{i,j}} \left[N_r \Lambda_{i,j}\right]^{m_{i,j}}}{m_{i,j}!}, 
    \label{eq: number_of_photons_poissons} 
\end{equation}
where $\Lambda_{i,j} = \eta \alpha_{i,j} S + B_{i,j}$. However, in cases of multiple detections, only the timestamp corresponding to the first photon is recorded—a phenomenon known as first-photon behavior. Note that since the scene is dynamic from frame to frame, the underlying photon arrival distribution at pixel $(i,j)$ varies over time.

Given that at least one photon is registered during each exposure time (i.e., $M > 0$), the timestamp distribution of the first photon detected at each pixel $(i,j)$ can be modeled as a mixture distribution, as follows \cite{Shin_2015_3D, Chan_2024_CVPR}:
\begin{equation} 
    \begin{split} 
    p(t_{i,j}) = \frac{\eta \alpha_{i,j} S}{\eta \alpha_{i,j} S + B_{i,j}} \left( \frac{s\left(t_{i,j} - \frac{2z_{i,j}}{c} - t_\text{jit}\right)}{S} \right) + \\ \frac{B_{i,j}}{\eta \alpha_{i,j} S + B_{i,j}} \left( \frac{1}{t_r} \right), 
    \label{eq: timestamp_distributions}
    \end{split} 
\end{equation}
where we model timing jitter as $t_\text{jit} \sim \mathcal{N}(0, \sigma_j)$, a temporal error interval, caused by the temporal response of the SPAD setup. Without loss of generality, we assume that the laser pulse $s(t)$ is a Gaussian pulse $\mathcal{N} \left( 0, \sigma_t \right)$, yielding $S = \int_{-t_r/2}^{t_r/2} s(t) dt = 1$. Therefore, the distribution can be further simplified to:
\begin{equation} 
\begin{split} 
    p(t_{i,j}) = \frac{\eta \alpha_{i,j}}{\eta \alpha_{i,j} + B_{i,j}} \left( s\left(t_{i,j} - \frac{2z_{i,j}}{c} - t_{\text{jit}}\right)\right) + \\ \frac{B_{i,j}}{\eta \alpha_{i,j} + B_{i,j}} \left( \frac{1}{t_r} \right).
    \end{split} 
    \label{eq: simplified_timestamp_distribution}
\end{equation}

To calculate the reflectivity of each pixel $\alpha_{i,j}$, i.e., the average number of photons per pixel, we use:
\begin{equation} 
    \alpha_{i,j} = \frac{E_0}{hc / \lambda} \cdot \frac{ 10^{-\alpha_{\text{atm}} \cdot 2R}  \ \Gamma_{i,j}}{8f_{\text{\#}}^2} \cdot \frac{W_p H_p}{A_{\text{illum}}},
    \label{eq: find_alpha}
\end{equation}
and the average number of photons from the background $B^{\text{bck}}_{i,j}$ and the dark current $B^{\text{dc}}$ are modeled, respectively, as:
\begin{equation} 
    B^{\text{bck}}_{i,j} = \frac{W^{\text{bck}}}{hc / \lambda} \cdot \frac{ 10^{-\alpha_\text{atm} \cdot R} \ \Gamma_{i,j}}{8f_{\text{\#}}^2}  \cdot  W_p H_p \cdot t_r, 
\end{equation}
\begin{equation} 
    B^{\text{dc}} = C^{\text{dc}} t_r,
\end{equation}
where $B_{i,j} = \eta B^{\text{bck}}_{i,j} + B^{\text{dc}} $, and $A_{\text{illum}}$ indicates the area of illumination by the source. In \cref{eq: find_alpha}, $E_0$ denotes the energy of the pulse, while $W_p$ and $H_p$ represent the width and height of a SPAD pixel. The variables $R$ and $\alpha_{atm}$ correspond to the range and atmospheric attenuation, respectively. The above equations inherently follow the models introduced in \cite{Scholes_2023_Fundamental}. All the notations and the corresponding parameter values used for this simulation are listed in \cref{tab: splidar_simulation_parameters}. 

\begin{figure*}[h] 
\centering 
    \includegraphics[width=0.97\linewidth]{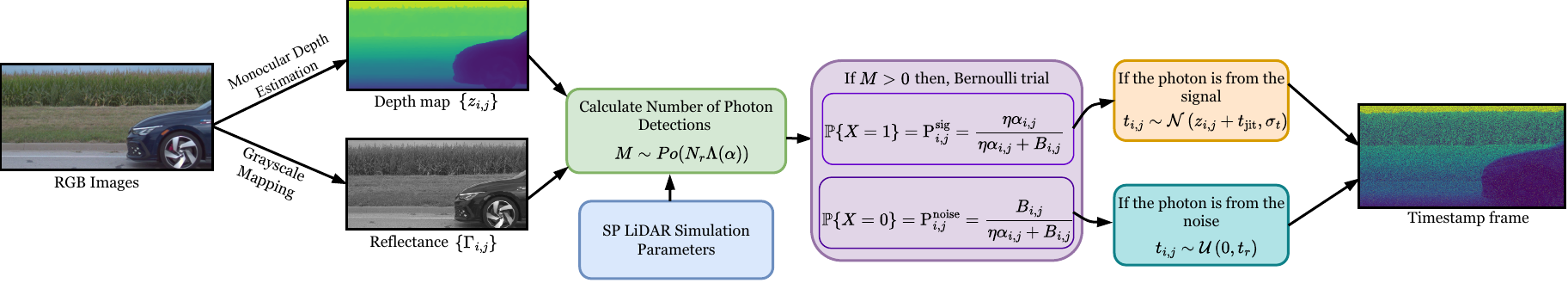} \caption{An overview of the single photon LiDAR timestamp simulation process demonstrates the key components of the pipeline. First, the number of photon detections at each pixel $(i,j)$ is obtained using a Poisson distribution. If there is at least one detection, a Bernoulli trial is then carried out to determine whether the detected photon originates from the signal or the noise. Once we sample from the corresponding distributions for all pixels, we can construct the timestamp frame.} 
    \label{fig: sp_lidar_simulation} 
\end{figure*}

\begin{table}[t]
\centering
\caption{SP-LiDAR Simulation Parameters}
\begin{tabular*}{0.5\textwidth}{@{\extracolsep{\fill}} c c c}
\toprule
Parameter       & Symbol & Value \\ \midrule
Dark counts               & $C^{\text{dc}}$ & $126$ Hz \\
Planck constant            & $h$ & $6.626 \times 10^{-34}$ m$^2$kgs$^{-1}$ \\
Wavelength                & $\lambda$       & $671$ nm \\
Attenuation coefficient   & $\alpha_{\text{atm}}$& $0.7 \text{dB km}^{-1}$ \\
f-number                  & $f_{\text{\#}}$ & $2.0$ \\
Reflectance               & $\Gamma_{i,j}$        & [$0.0$, $1.0$] \\
Height of an effective pixel &  $H_p$       & 9.2 $\mu$m \\
Width of an effective pixel  &  $W_p$       & 9.2 $\mu$m \\
Exposure time            & $t_{\text{exp}}$        & $1000 \ \mu$s  \\
Jitter variation          & $\sigma_j$             & $220$ ps \\
Range                    & $R$             & $30$ m \\
Pulse width              & $\sigma_t$       & $1$ ns \\
Energy per pulse         & $E_0$           & $1.219$ nJ \\
Background radiation     & $W^{\text{bck}}$& $0.0002$ W \\
Repetition rate          & $1/t_r$           & $2.25$ MHz \\
Depth variation          & $z_{i,j}$             & [$2$ m, $60$ m] \\ 
Efficiency of the sensor & $\eta$             & $0.18$ \\ 
Speed of Light           & $c$             & $3 \times 10^8 \text{ms}^{-1}$ \\ 
\bottomrule
\end{tabular*}
\label{tab: splidar_simulation_parameters}
\end{table}

To obtain high-speed depth maps, we utilized an RGB $2000$-FPS video dataset \cite{chennuri_2024_quiver}, which was subsequently processed using the monocular depth estimation algorithm \textit{Depth Anything V2} \cite{yang2024depthv2} to generate ground truth depth maps. Using pre-trained models, we could only acquire relative depth variations. Therefore, these depth map values were converted into absolute depth values, denoted as ${z_{i,j}}$. For reflectance ${\Gamma_{i,j}}$, we used the grayscale values of the RGB images, a common practice in SP-LiDAR literature.

Once all parameters were obtained, we sampled timestamps from the PDF provided in \cref{eq: simplified_timestamp_distribution}. Since the corresponding process is a mixture of Gaussian and Uniform distributions, we could separate the PDF to determine whether a detected photon originated from the signal or noise based on the outcomes of a Bernoulli distribution defined as follows:
\begin{equation}
\mathbb{P}\{X = x\} =
\begin{cases}
\mathrm{P}^{\text{sig}}_{i,j} = \dfrac{\eta \alpha_{i,j}}{\eta \alpha_{i,j} + B_{i,j}}, & \text{if } x = 1, \\
\mathrm{P}^{\text{noise}}_{i,j} = \dfrac{B_{i,j}}{\eta \alpha_{i,j} + B_{i,j}}, & \text{if } x = 0,
\end{cases}
\end{equation}
where $X=1$ denotes the event that the detected photon originates from the signal, with corresponding probability $\mathrm{P}^{\text{sig}}_{i,j}$, and ${X=0}$ denotes the event that the detected photon originates from noise, with corresponding probability $\mathrm{P}^{\text{noise}}_{i,j}$. If the outcome of the Bernoulli trial is signal (i.e., if ${X=1}$), the timestamp is sampled from the corresponding Gaussian distribution. Otherwise (i.e., if ${X=0}$), the timestamp is sampled from the uniform distribution $\mathcal{U}(0, t_r)$. This sampling mechanism is applied to all pixels in each frame, across all frames in each video. The overall simulation process is illustrated in \cref{fig: sp_lidar_simulation}.

\begin{figure*}[t] 
\centering 
    \includegraphics[width=0.95\linewidth]{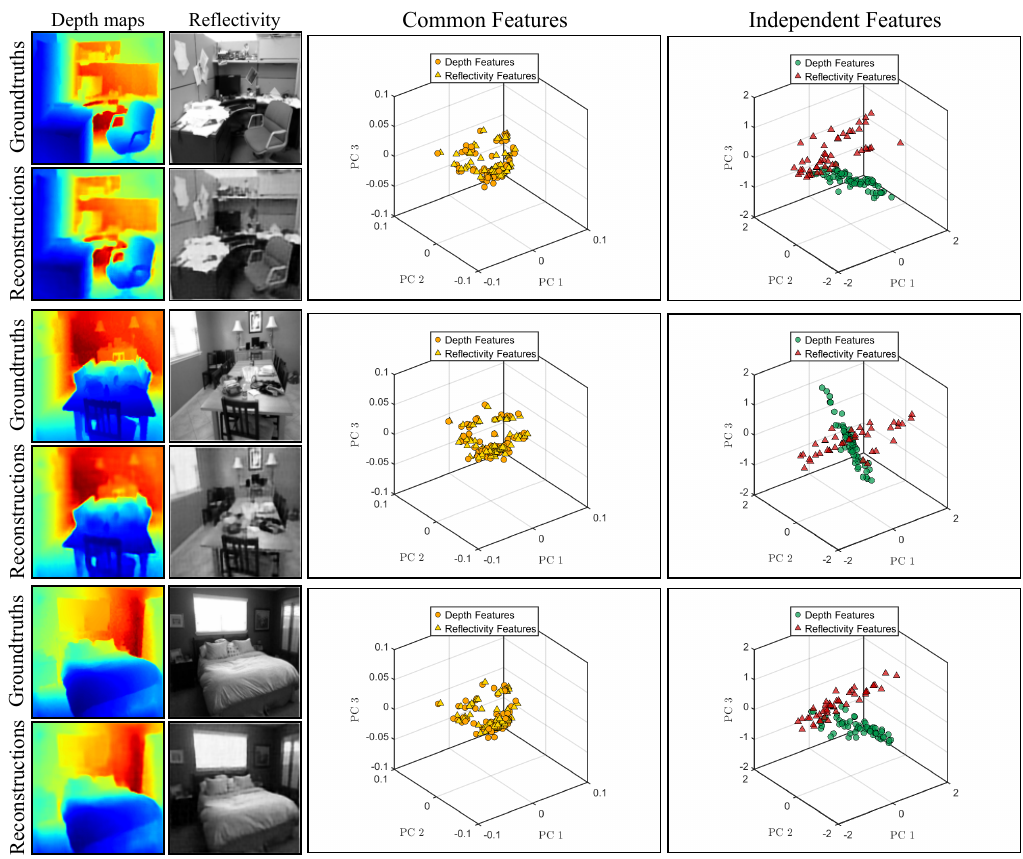} \caption{The feature distribution in the low-dimensional space for four scenes shows that the common features are co-located, while the remaining features are dispersed. ``PC $1$'', ``PC $2$'' and ``PC $3$'' on the axes represent the $1^{\text{th}}$, $2^{\text{nd}}$ and $3^{\text{rd}}$ principal components.} 
    \label{fig: feature_distribution} 
\end{figure*}

\section{Visualization of Common Features}
\label{sec: visualization_features}

In this section, we provide the visualization of the common features from the experiment conducted in 
Sec.~\textcolor{blue}{III-C}, focusing on the latent space of both reflectivity and depth. The input images for depth and reflectivity have a size of $128 \times 128 \times 1$. The feature map size is gradually reduced through a series of convolutional layers, resulting in a latent feature map of $8 \times 8 \times 128$, corresponding to a $2\times$ compression ratio for each branch. One-third of the latent feature map, i.e., a feature map of size $8 \times 8 \times 43$, is trained to extract the common features shared by both depth and reflectivity.

To visualize these common features in a low-dimensional space, we use Principal Component Analysis (PCA), a dimensionality reduction technique that linearly transforms the data into a new coordinate system, making it easier to identify the principal components (PCs) that capture the most significant variations  \cite{cadima_2016_pca}. Using PCA, the original $128 \times 43$ unrolled common feature maps of reflectivity and depth, denoted as $\mathcal{H}_{\text{lsFeat}} = [\mathcal{R}_{\text{lsFeat}}, \mathcal{D}_{\text{lsFeat}}]$, are projected onto a $128 \times 3$ low-dimensional feature space. Mathematically, the projection is formulated as
\begin{equation} 
\begin{split} 
    \mathcal{H}^\text{pca}_{\text{lsFeat}} = \mathcal{H}^c_{\text{lsFeat}} W,
    \end{split} 
    \label{eq: pca_formulation}
\end{equation}
where $W  = [v_1 \ v_2 \cdots v_k] \in \mathbb{R}^{p \times k}$ is the projection matrix, constructed from the eigenvectors corresponding to the largest $k$ eigenvalues of the covariance matrix of $\mathcal{H}^c_{\text{lsFeat}} \in \mathbb{R}^{n \times p}$, the centered latent feature maps. Here, $p = 43$ and $n = 128$ denote the number of features and the number of samples, respectively, while $k = 3$ represents the number of PCs. The resulting transformed feature map is $\mathcal{H}^\text{pca}_{\text{lsFeat}} \in \mathbb{R}^{n \times k}$. Similarly, to showcase independent latent features of depth and reflectivity, the same procedure is followed.

\cref{fig: feature_distribution} illustrates the distribution of common and independent features in the low-dimensional space. Across the feature distributions of the three scenarios, we observe that the common features are co-located, while the independent features are clearly separated. These results further support the presence of shared features between the depth and reflectivity inputs.

\bibliographystyle{IEEEtran}
\bibliography{main}

\begin{thebibliography}{10}
\providecommand{\url}[1]{#1}
\csname url@samestyle\endcsname
\providecommand{\newblock}{\relax}
\providecommand{\bibinfo}[2]{#2}
\providecommand{\BIBentrySTDinterwordspacing}{\spaceskip=0pt\relax}
\providecommand{\BIBentryALTinterwordstretchfactor}{4}
\providecommand{\BIBentryALTinterwordspacing}{\spaceskip=\fontdimen2\font plus
\BIBentryALTinterwordstretchfactor\fontdimen3\font minus \fontdimen4\font\relax}
\providecommand{\BIBforeignlanguage}[2]{{%
\expandafter\ifx\csname l@#1\endcsname\relax
\typeout{** WARNING: IEEEtran.bst: No hyphenation pattern has been}%
\typeout{** loaded for the language `#1'. Using the pattern for}%
\typeout{** the default language instead.}%
\else
\language=\csname l@#1\endcsname
\fi
#2}}
\providecommand{\BIBdecl}{\relax}
\BIBdecl

\bibitem{boretti_2024_prespective_lidar}
A.~Boretti, ``\href{https://doi.org/10.1002/mop.33918 }{A Perspective on Single-Photon {LiDAR} Systems},'' \emph{Microwave and Optical Technology Letters}, vol.~66, no.~1, p. e33918, 2024.

\bibitem{Morimoto_2020_SPAD}
K.~Morimoto, A.~Ardelean, M.-L. Wu, A.~C. Ulku, I.~M. Antolovic, C.~Bruschini, and E.~Charbon, ``\href{https://doi.org/10.1364/OPTICA.386574 }{Megapixel Time-gated {SPAD} Image Sensor for 2{D} and 3{D} Imaging Applications},'' \emph{Optica}, vol.~7, no.~4, pp. 346--354, 2020.

\bibitem{Li_2023_SPAD}
Z.~Li, H.~Pan, G.~Shen, D.~Zhai, W.~Zhang, L.~Yang, and G.~Wu, ``\href{https://doi.org/10.1016/j.optlastec.2022.108749 }{Single-photon {LiDAR} for Canopy Detection with a Multi-Channel {Si} {SPAD} at 1064nm},'' \emph{Optics \& Laser Technology}, vol. 157, p. 108749, 2023.

\bibitem{Rapp_2020_SPM}
J.~Rapp, J.~Tachella, Y.~Altmann, S.~McLaughlin, and V.~K. Goyal, ``\href{https://doi.org/10.1109/MSP.2020.2983772 }{Advances in Single-Photon Lidar for Autonomous Vehicles: Working Principles, Challenges, and Recent Advances},'' \emph{IEEE Signal Processing Magazine}, vol.~37, no.~4, pp. 62--71, 2020.

\bibitem{McCarthy_2013_kmrange}
A.~McCarthy, N.~J. Krichel, N.~R. Gemmell, X.~Ren, M.~G. Tanner, S.~N. Dorenbos, V.~Zwiller, R.~H. Hadfield, and G.~S. Buller, ``\href{https://doi.org/10.1364/OE.21.008904 }{Kilometer-range, High Resolution Depth Imaging via 1560 nm Wavelength Single-Photon Detection},'' \emph{Opt. Express}, vol.~21, no.~7, pp. 8904--8915, 2013.

\bibitem{mora_martin_21_object_detection}
G.~Mora-Mart\'{i}n, A.~Turpin, A.~Ruget, A.~Halimi, R.~Henderson, J.~Leach, and I.~Gyongy, ``\href{https://doi.org/10.1364/OE.435619 }{High-speed Object Detection with a Single-photon Time-of-Flight image sensor},'' \emph{Opt. Express}, vol.~29, no.~21, pp. 33\,184--33\,196, 2021.

\bibitem{Li_2020_LiDARReview}
Y.~Li and J.~Ibanez-Guzman, ``\href{https://doi.org/10.1109/MSP.2020.2973615 }{{LiDAR} for Autonomous Driving: The Principles, Challenges, and Trends for Automotive {LiDAR} and Perception Systems},'' \emph{IEEE Signal Processing Magazine}, vol.~37, no.~4, pp. 50--61, 2020.

\bibitem{Becker_2005_TCSPC}
W.~Becker, \emph{\href{https://doi.org/10.1007/3-540-28882-1 }{Advanced Time-Correlated Single Photon Counting Techniques}}, 2nd~ed.\hskip 1em plus 0.5em minus 0.4em\relax Springer, 2005.

\bibitem{Charbon_2013_spad-based}
E.~Charbon, M.~Fishburn, R.~Walker, R.~K. Henderson, and C.~Niclass, \emph{\href{https://doi.org/10.1007/978-3-642-27523-4_2 }{{SPAD}-Based Sensors in TOF Range-Imaging Cameras}}.\hskip 1em plus 0.5em minus 0.4em\relax Springer, 2013, pp. 11--38.

\bibitem{Edoardo_2019_modelling}
P.~Padmanabhan, C.~Zhang, and E.~Charbon, ``\href{https://doi.org/10.3390/s19245464 }{Modeling and Analysis of a Direct Time-of-Flight Sensor Architecture for {LiDAR} Applications},'' \emph{MDPI, Sensors}, vol.~19, no.~24, p. 5464, 2019.

\bibitem{Torben_2023_spad_simulation}
T.~Neumann and F.~Kallage, ``\href{https://doi.org/10.1109/JSEN.2023.3275269 }{Simulation of a Direct Time-of-Flight {LiDAR}-System},'' \emph{IEEE Sensors Journal}, vol.~23, no.~13, pp. 14\,245--14\,252, 2023.

\bibitem{rapp_2017_unmixing}
J.~Rapp and V.~K. Goyal, ``\href{https://doi.org/10.1109/TCI.2017.2706028 }{A Few Photons Among Many: Unmixing Signal and Noise for Photon-Efficient Active Imaging},'' \emph{IEEE Transactions on Computational Imaging (TCI)}, vol.~3, no.~3, pp. 445--459, 2017.

\bibitem{Shin_2015_3D}
D.~Shin, A.~Kirmani, V.~K. Goyal, and J.~H. Shapiro, ``\href{https://doi.org/10.1109/TCI.2015.2453093 }{Photon-Efficient Computational 3-{D} and Reflectivity Imaging With Single-Photon Detectors},'' \emph{IEEE Transactions on Computational Imaging (TCI)}, vol.~1, no.~2, pp. 112--125, 2015.

\bibitem{peng_2020_non_local}
J.~Peng, Z.~Xiong, X.~Huang, Z.-P. Li, D.~Liu, and F.~Xu, ``\href{https://doi.org/10.1007/978-3-030-58539-6_14 }{Photon-Efficient 3{D} Imaging with A Non-local Neural Network},'' in \emph{European Conference on Computer Vision (ECCV)}, 2020, pp. 225--241.

\bibitem{Lindell_2018_SIGGRAPH}
D.~B. Lindell, M.~O'Toole, and G.~Wetzstein, ``\href{https://doi.org/10.1145/3197517.3201316}{Single-Photon 3{D} Imaging with Deep Sensor Fusion},'' \emph{ACM Trans. Graph.}, vol.~37, no.~4, pp. 113.1--113.12, 2018.

\bibitem{peng_2023_boosting}
J.~Peng, Z.~Xiong, H.~Tan, X.~Huang, Z.-P. Li, and F.~Xu, ``\href{https://doi.org/10.1109/TPAMI.2022.3200745 }{Boosting Photon-Efficient Image Reconstruction With A Unified Deep Neural Network},'' \emph{IEEE Transactions on Pattern Analysis and Machine Intelligence (TPAMI)}, vol.~45, no.~4, pp. 4180--4197, 2023.

\bibitem{under_water_2023_Maccarone}
A.~Maccarone, K.~Drummond, A.~McCarthy, U.~K. Steinlehner, J.~Tachella, D.~A. Garcia, A.~Pawlikowska, R.~A. Lamb, R.~K. Henderson, S.~McLaughlin, Y.~Altmann, and G.~S. Buller, ``\href{https://doi.org/10.1364/OE.487129 }{Submerged Single-Photon {LiDAR} Imaging Sensor Used for Real-time 3{D} Scene Reconstruction in Scattering Underwater Environments},'' \emph{Opt. Express}, vol.~31, no.~10, pp. 16\,690--16\,708, 2023.

\bibitem{halimi_2017_joint_underwater}
A.~Halimi, A.~Maccarone, A.~McCarthy, S.~McLaughlin, and G.~S. Buller, ``\href{https://doi.org/10.1109/TCI.2017.2669867 }{Object Depth Profile and Reflectivity Restoration From Sparse Single-Photon Data Acquired in Underwater Environments},'' \emph{IEEE Transactions on Computational Imaging (TCI)}, vol.~3, no.~3, pp. 472--484, 2017.

\bibitem{fog_2022_zhang}
Y.~Zhang, S.~Li, J.~Sun, X.~Zhang, D.~Liu, X.~Zhou, H.~Li, and Y.~Hou, ``\href{https://doi.org/10.1364/OE.464297 }{Three-dimensional Single-Photon Imaging through Realistic Fog in an Outdoor Environment During the Day},'' \emph{Opt. Express}, vol.~30, no.~19, pp. 34\,497--34\,509, 2022.

\bibitem{Shin_2016_multidepth}
D.~Shin, F.~Xu, F.~N.~C. Wong, J.~H. Shapiro, and V.~K. Goyal, ``\href{https://doi.org/10.1364/OE.24.001873 }{Computational Multi-depth Single-Photon Imaging},'' \emph{Opt. Express}, vol.~24, no.~3, pp. 1873--1888, 2016.

\bibitem{altmann_2020_individual}
Y.~Altmann, S.~McLaughlin, and M.~E. Davies, ``\href{https://doi.org/10.1109/TIP.2019.2952008 }{Fast Online {3D} Reconstruction of Dynamic Scenes From Individual Single-Photon Detection Events},'' \emph{IEEE Transactions on Image Processing (TIP)}, vol.~29, pp. 2666--2675, 2020.

\bibitem{kirmani_first-photon_2014}
A.~Kirmani, D.~Venkatraman, D.~Shin, A.~Colaço, F.~N.~C. Wong, J.~H. Shapiro, and V.~K. Goyal, ``\href{https://doi.org/10.1126/science.1246775 }{First-Photon Imaging},'' \emph{Science}, vol. 343, no. 6166, pp. 58--61, 2014.

\bibitem{Altmann_2016_LiDAR}
Y.~Altmann, X.~Ren, A.~McCarthy, G.~S. Buller, and S.~McLaughlin, ``\href{https://doi.org/10.1109/TIP.2016.2526784}{LiDAR Waveform-Based Analysis of Depth Images Constructed Using Sparse Single-Photon Data},'' \emph{IEEE Transactions on Image Processing (TIP)}, vol.~25, no.~5, pp. 1935--1946, 2016.

\bibitem{Tachella_2019_complex}
J.~Tachella, Y.~Altmann, X.~Ren, A.~McCarthy, G.~S. Buller, S.~McLaughlin, and J.-Y. Tourneret, ``\href{https://doi.org/10.1137/18M1183972 }{{B}ayesian 3{D} Reconstruction of Complex Scenes from Single-Photon {LiDAR} Data},'' \emph{SIAM Journal on Imaging Sciences}, vol.~12, no.~1, pp. 521--550, 2019.

\bibitem{Altman_2016_Bayesian}
Y.~Altmann, X.~Ren, A.~McCarthy, G.~S. Buller, and S.~McLaughlin, ``\href{https://doi.org/10.1109/TCI.2016.2618323 }{Robust {B}ayesian Target Detection Algorithm for Depth Imaging From Sparse Single-Photon Data},'' \emph{IEEE Transactions on Computational Imaging (TCI)}, vol.~2, no.~4, pp. 456--467, 2016.

\bibitem{lee_caspi_2023}
J.~Lee, A.~Ingle, J.~V. Chacko, K.~W. Eliceiri, and M.~Gupta, ``\href{https://doi.org/10.1038/s41467-023-38893-9 }{{CASPI}: Collaborative Photon Processing for Active Single-Photon Imaging},'' \emph{Nature Communications}, vol.~14, no.~1, p. 3158, 2023.

\bibitem{Rapp_2019_Dead}
J.~Rapp, Y.~Ma, R.~M.~A. Dawson, and V.~K. Goyal, ``\href{https://doi.org/10.1109/TSP.2019.2914891 }{Dead Time Compensation for High-Flux Ranging},'' \emph{IEEE Transactions on Signal Processing}, vol.~67, no.~13, pp. 3471--3486, 2019.

\bibitem{Pediredla_2018_pileup}
A.~K. Pediredla, A.~C. Sankaranarayanan, M.~Buttafava, A.~Tosi, and A.~Veeraraghavan, ``\href{https://doi.org/10.48550/arXiv.1806.07437 }{Signal Processing Based Pile-up Compensation for Gated Single-Photon Avalanche Diodes},'' \emph{arXiv preprint arXiv: 1806.07437}, 2018.

\bibitem{Gupta_2019_Flooded}
A.~Gupta, A.~Ingle, A.~Velten, and M.~Gupta, ``\href{https://doi.org/10.1109/CVPR.2019.00693 }{Photon-Flooded Single-Photon 3{D} Cameras},'' in \emph{IEEE/CVF Conference on Computer Vision and Pattern Recognition (CVPR)}, 2019, pp. 6763--6772.

\bibitem{zhang_2024_em}
W.~Zhang, H.~K. Weerasooriya, P.~Chennuri, and S.~H. Chan, ``\href{https://doi.org/10.1109/MMSP61759.2024.10743884 }{Parametric Modeling and Estimation of Photon Registrations for {3D} Imaging},'' in \emph{IEEE 26th International Workshop on Multimedia Signal Processing (MMSP)}, 2024, pp. 1--6.

\bibitem{Bar-David_1969}
I.~Bar-David, ``\href{https://doi.org/10.1109/TIT.1969.1054238}{Communication under the {P}oisson regime},'' \emph{IEEE Transactions on Information Theory}, vol.~15, no.~1, pp. 31--37, 1969.

\bibitem{Chan_2024_CVPR}
S.~H. Chan, H.~K. Weerasooriya, W.~Zhang, P.~Abshire, I.~Gyongy, and R.~K. Henderson, ``\href{https://doi.org/10.1109/CVPR52733.2024.02391 }{Resolution Limit of Single-Photon {LiDAR}},'' in \emph{IEEE/CVF Conference on Computer Vision and Pattern Recognition (CVPR)}, 2024, pp. 25\,307--25\,316.

\bibitem{Yau_2024_MMSP}
W.~C. Yau, W.~Zhang, H.~K. Weerasooriya, and S.~H. Chan, ``\href{https://doi.org/10.1109/MMSP61759.2024.10743397 }{Analysis and Improvement of Rank-Ordered Mean Algorithm in Single-Photon {LiDAR}},'' in \emph{IEEE 26th International Workshop on Multimedia Signal Processing (MMSP)}, 2024, pp. 1--6.

\bibitem{yao_2022_sparsity}
G.~Yao, Y.~Chen, C.~Jiang, Y.~Xuan, X.~Hu, Y.~Liu, and Y.~Pan, ``\href{https://doi.org/10.1364/OE.471610 }{Dynamic Single-Photon 3{D} Imaging with a Sparsity-Based Neural Network},'' \emph{Opt. Express}, vol.~30, no.~21, pp. 37\,323--37\,340, 2022.

\bibitem{zhao_2022_edge_enhancement}
X.~Zhao, X.~Jiang, A.~Han, T.~Mao, W.~He, and Q.~Chen, ``\href{https://doi.org/10.1364/OE.446369 }{Photon-efficient 3{D} Reconstruction Employing a Edge Enhancement Method},'' \emph{Opt. Express}, vol.~30, no.~2, pp. 1555--1569, 2022.

\bibitem{zang_2021_non_fusion}
Z.~Zang, D.~Xiao, and D.~D.-U. Li, ``\href{https://doi.org/10.1364/OE.425917 }{Non-fusion Time-resolved Depth Image Reconstruction using a Highly Efficient Neural Network Architecture},'' \emph{Opt. Express}, vol.~29, no.~13, pp. 19\,278--19\,291, 2021.

\bibitem{ruget_2021_multi_feature}
A.~Ruget, S.~McLaughlin, R.~K. Henderson, I.~Gyongy, A.~Halimi, and J.~Leach, ``\href{https://doi.org/10.1364/OE.415563 }{Robust Super-Resolution Depth Imaging via a Multi-feature Fusion Deep Network},'' \emph{Opt. Express}, vol.~29, no.~8, pp. 11\,917--11\,937, 2021.

\bibitem{sun_2020_monocular_fusion}
Z.~Sun, D.~B. Lindell, O.~Solgaard, and G.~Wetzstein, ``\href{https://doi.org/10.1364/OE.392386 }{{SPAD}net: Deep {RGB-SPAD} Sensor Fusion Assisted by Monocular Depth Estimation},'' \emph{Opt. Express}, vol.~28, no.~10, pp. 14\,948--14\,962, 2020.

\bibitem{Altman_2017_Binary}
Y.~Altmann, R.~Aspden, M.~Padgett, and S.~McLaughlin, ``\href{https://doi.org/10.1109/TCI.2017.2703900 }{A {B}ayesian Approach to Denoising of Single-Photon Binary Images},'' \emph{IEEE Transactions on Computational Imaging (TCI)}, vol.~3, no.~3, pp. 460--471, 2017.

\bibitem{fossum_2016_quanta}
E.~R. Fossum, J.~Ma, S.~Masoodian, L.~Anzagira, and R.~Zizza, ``\href{https://doi.org//10.3390/s16081260 }{The Quanta Image Sensor: Every Photon Counts},'' \emph{MDPI, Sensors}, vol.~16, no.~8, p. 1260, 2016.

\bibitem{ma_2022_review}
J.~Ma, S.~Chan, and E.~R. Fossum, ``\href{https://doi.org/10.1109/TED.2022.3166716 }{Review of Quanta Image Sensors for Ultralow-Light Imaging},'' \emph{IEEE Transactions on Electron Devices}, vol.~69, no.~6, pp. 2824--2839, 2022.

\bibitem{chennuri_2024_quiver}
P.~Chennuri, Y.~Chi, E.~Jiang, G.~M.~D. Godaliyadda, A.~Gnanasambandam, H.~R. Sheikh, I.~Gyongy, and S.~H. Chan, ``\href{https://doi.org/10.1007/978-3-031-73661-2_9 }{Quanta Video Restoration},'' in \emph{European Conference on Computer Vision (ECCV)}, 2024, pp. 152--171.

\bibitem{Ma_2020_QBP}
S.~Ma, S.~Gupta, A.~C. Ulku, C.~Brushini, E.~Charbon, and M.~Gupta, ``\href{https://doi.org/10.1145/3386569.3392470 }{Quanta Burst Photography},'' \emph{ACM Transactions on Graphics (TOG)}, vol.~39, no.~4, pp. 79:1 -- 79:16, 2020.

\bibitem{Liu_2024_bit2bit}
Y.~Liu, A.~Krull, H.~Basevi, A.~Leonardis, and M.~Jenkins, ``\href{https://doi.org/10.48550/arXiv.2410.23247}{Bit2Bit: 1-bit Quanta Video Reconstruction via Self-Supervised Photon Prediction},'' in \emph{Advances in Neural Information Processing Systems (NeurIPS)}, 2024, pp. 88\,443--88\,485.

\bibitem{chi_2020_dynamic}
Y.~Chi, A.~Gnanasambandam, V.~Koltun, and S.~H. Chan, ``\href{https://doi.org/10.1007/978-3-030-58589-1_8 }{Dynamic Low-Light Imaging with Quanta Image Sensors},'' in \emph{European Conference on Computer Vision (ECCV)}, 2020, pp. 122--138.

\bibitem{chan_2016_images}
S.~H. Chan, O.~A. Elgendy, and X.~Wang, ``\href{https://doi.org/10.3390/s16111961 }{Images from Bits: Non-iterative Image Reconstruction for Quanta Image Sensors},'' \emph{MDPI, Sensors}, vol.~16, no.~11, p. 1961, 2016.

\bibitem{choi_2018_qis}
J.~H. Choi, O.~A. Elgendy, and S.~H. Chan, ``\href{https://doi.org/10.1109/ICASSP.2018.8461685 }{Image Reconstruction for Quanta Image Sensors using Deep Neural Networks},'' in \emph{IEEE International Conference on Acoustics, Speech and Signal Processing (ICASSP)}, 2018, pp. 6543--6547.

\bibitem{zhang_2024_streamingqis}
T.~Zhang, M.~Dutson, V.~Boominathan, M.~Gupta, and A.~Veeraraghavan, ``\href{https://doi.org/10.1109/TPAMI.2024.3501154 }{Streaming Quanta Sensors for Online, High-Performance Imaging and Vision},'' \emph{IEEE Transactions on Pattern Analysis and Machine Intelligence (TPAMI)}, vol.~47, no.~3, pp. 1564--1577, 2025.

\bibitem{vivek_2024_detection}
R.~Kitichotkul, J.~Rapp, and V.~K. Goyal, ``\href{https://doi.org/10.1109/JSTQE.2023.3333834 }{The Role of Detection Times in Reflectivity Estimation with Single-Photon {LiDAR}},'' \emph{IEEE Journal of Selected Topics in Quantum Electronics}, vol.~30, no.~1, pp. 1--14, 2024.

\bibitem{drummond_joint_estimate}
K.~Drummond, S.~McLaughlin, Y.~Altmann, A.~Pawlikowska, and R.~Lamb, ``\href{https://doi.org/10.1109/SSPD51364.2021.9541522 }{Joint Surface Detection and Depth Estimation from Single-Photon {LiDAR} Data using Ensemble Estimators},'' in \emph{Sensor Signal Processing for Defence Conference (SSPD)}, 2021, pp. 1--5.

\bibitem{halimi_2016_joint_admm}
A.~Halimi, Y.~Altmann, A.~McCarthy, X.~Ren, R.~Tobin, G.~S. Buller, and S.~McLaughlin, ``\href{https://doi.org/10.1109/EUSIPCO.2016.7760215 }{Restoration of Intensity and Depth Images Constructed using Sparse Single-Photon Data},'' in \emph{European Signal Processing Conference (EUSIPCO)}, 2016, pp. 86--90.

\bibitem{gyongy_2020_high_speed}
I.~Gyongy, S.~W. Hutchings, A.~Halimi, M.~Tyler, S.~Chan, F.~Zhu, S.~McLaughlin, R.~K. Henderson, and J.~Leach, ``\href{https://doi.org/10.1364/OPTICA.390099 }{High-speed 3{D} Sensing via Hybrid-mode Imaging and Guided Upsampling},'' \emph{Optica}, no.~10, pp. 1253--1260, 2020.

\bibitem{Snyder_1991_book}
D.~L. Snyder and M.~I. Miller, \emph{\href{https://doi.org/10.1007/978-1-4612-3166-0}{Random Point Processes in Time and Space}}.\hskip 1em plus 0.5em minus 0.4em\relax Springer, 1991.

\bibitem{Silberman_ECCV12_dataset}
N.~Silberman, D.~Hoiem, P.~Kohli, and R.~Fergus, ``\href{https://doi.org/10.1007/978-3-642-33715-4_54 }{Indoor Segmentation and Support Inference from {RGBD} Images},'' in \emph{European Conference on Computer Vision (ECCV)}, 2012, pp. 746--760.

\bibitem{zhang_2018_depth_completion}
Y.~Zhang and T.~Funkhouser, ``\href{https://doi.org/10.1109/CVPR.2018.00026 }{Deep Depth Completion of a Single {RGB-D} Image},'' in \emph{IEEE/CVF Conference on Computer Vision and Pattern Recognition (CVPR)}, 2018, pp. 175--185.

\bibitem{zhang2023completionformer}
Y.~Zhang, X.~Guo, M.~Poggi, Z.~Zhu, G.~Huang, and S.~Mattoccia, ``\href{https://doi.org/10.1109/CVPR52729.2023.01777 }{Completionformer: Depth Completion with Convolutions and Vision Transformers},'' in \emph{IEEE/CVF Conference on Computer Vision and Pattern Recognition (CVPR)}, 2023, pp. 18\,527--18\,536.

\bibitem{Steve_2016_computer_graphics}
S.~Marschner and P.~Shirley, \emph{\href{https://dl.acm.org/doi/10.5555/2872329 }{Fundamentals of Computer Graphics}}, 4th~ed.\hskip 1em plus 0.5em minus 0.4em\relax A. K. Peters, Ltd., 2016.

\bibitem{doCarmo_2016_DifferentialGeometry}
M.~P. do~Carmo, \emph{\href{https://doi.org/10.1007/978-3-319-39799-3 }{Differential Geometry of Curves and Surfaces}}, 2nd~ed.\hskip 1em plus 0.5em minus 0.4em\relax Dover Publications, 2016.

\bibitem{woo_2018_cbam}
S.~Woo, J.~Park, J.-Y. Lee, and I.~S. Kweon, ``\href{https://doi.org/10.1007/978-3-030-01234-2_1 }{{CBAM}: Convolutional Block Attention Module},'' in \emph{European Conference on Computer Vision (ECCV)}, 2018, pp. 3--19.

\bibitem{vaswani_2017_attention}
A.~Vaswani, N.~Shazeer, N.~Parmar, J.~Uszkoreit, L.~Jones, A.~N. Gomez, L.~u. Kaiser, and I.~Polosukhin, ``\href{ https://doi.org/10.48550/arXiv.1706.03762 }{Attention is All you Need},'' in \emph{Advances in Neural Information Processing Systems (NeurIPS)}, 2017, pp. 5998--6008.

\bibitem{zhou2022revisiting}
K.~Zhou, W.~Li, L.~Lu, X.~Han, and J.~Lu, ``\href{https://doi.org/10.1109/CVPR52688.2022.00596 }{Revisiting Temporal Alignment for Video Restoration},'' in \emph{IEEE/CVF Conference on Computer Vision and Pattern Recognition (CVPR)}, 2022, pp. 6053--6062.

\bibitem{tu2022maxvit}
Z.~Tu, H.~Talebi, H.~Zhang, F.~Yang, P.~Milanfar, A.~Bovik, and Y.~Li, ``\href{https://doi.org/10.1007/978-3-031-20053-3_27 }{Max{V}i{T}: Multi-axis Vision Transformer},'' in \emph{European Conference on Computer Vision (ECCV)}, 2022, pp. 459--479.

\bibitem{Henderson_2019_192x128}
R.~K. Henderson, N.~Johnston, F.~Mattioli Della~Rocca, H.~Chen, D.~Day-Uei~Li, G.~Hungerford, R.~Hirsch, D.~Mcloskey, P.~Yip, and D.~J.~S. Birch, ``\href{https://doi.org/10.1109/JSSC.2019.2905163 }{A $192\times128$ Time Correlated {SPAD} Image Sensor in 40-nm {CMOS} Technology},'' \emph{IEEE Journal of Solid-State Circuits}, vol.~54, no.~7, pp. 1907--1916, 2019.

\bibitem{Scholes_2023_Fundamental}
S.~Scholes, G.~Mora-Martin, F.~Zhu, I.~Gyongy, P.~Soan, and J.~Leach, ``\href{https://doi.org/10.1038/s41598-022-27012-1 }{Fundamental Limits to Depth Imaging with Single-photon Detector Array Sensors},'' \emph{Nature Scientific Reports}, vol.~13, no.~1, p. 176, 2023.

\bibitem{yang2024depthv2}
L.~Yang, B.~Kang, Z.~Huang, Z.~Zhao, X.~Xu, J.~Feng, and H.~Zhao, ``\href{ https://doi.org/10.48550/arXiv.2406.09414 }{Depth Anything V2},'' in \emph{Advances in Neural Information Processing Systems (NeurIPS)}, 2024, pp. 21\,875--21\,911.

\bibitem{Kingma_2015_adam}
D.~P. Kingma and J.~Ba, ``\href{https://doi.org/10.48550/arXiv.1412.6980}{Adam: {A} Method for Stochastic Optimization},'' in \emph{International Conference on Learning Representations, {(ICLR)}}, 2015.

\bibitem{Ji_2022_memdeblur}
B.~Ji and A.~Yao, ``\href{https://doi.org/10.1109/CVPR52688.2022.00196 }{Multi-Scale Memory-Based Video Deblurring},'' in \emph{IEEE/CVF Conference on Computer Vision and Pattern Recognition (CVPR)}, 2022, pp. 1919--1928.

\bibitem{zhao_2021_spike}
J.~Zhao, R.~Xiong, H.~Liu, J.~Zhang, and T.~Huang, ``\href{https://doi.org/10.1109/CVPR46437.2021.01182 }{{Spk2ImgNet}: Learning to Reconstruct Dynamic Scene from Continuous Spike Stream},'' in \emph{IEEE/CVF Conference on Computer Vision and Pattern Recognition (CVPR)}, 2021, pp. 11\,991--12\,000.

\bibitem{liang_2022_rvrt}
J.~Liang, Y.~Fan, X.~Xiang, R.~Ranjan, E.~Ilg, S.~Green, J.~Cao, K.~Zhang, R.~Timofte, and L.~V. Gool, ``\href{ https://doi.org/10.48550/arXiv.2206.02146 }{Recurrent Video Restoration Transformer with Guided Deformable Attention},'' in \emph{Advances in Neural Information Processing Systems (NeurIPS)}, 2022, pp. 378--393.

\bibitem{li_2022_flornn}
J.~Li, X.~Wu, Z.~Niu, and W.~Zuo, ``\href{https://doi.org/10.1007/978-3-031-19797-0_34 }{Unidirectional Video Denoising by Mimicking Backward Recurrent Modules with Look-Ahead Forward Ones},'' in \emph{European Conference on Computer Vision (ECCV)}, 2022, pp. 592--609.

\bibitem{cadima_2016_pca}
I.~T. Jolliffe and J.~Cadima, ``\href{https://doi.org/10.1098/rsta.2015.0202 }{Principal Component Analysis: A Review and Recent Developments},'' \emph{Philosophical Transactions of the Royal Society A: Mathematical, Physical and Engineering Sciences}, vol. 374, no. 2065, p. 20150202, 2016.

\end{thebibliography}



\end{document}